\documentclass[authoryear,final,times,11pt]{elsarticle}
\usepackage{lineno,hyperref}
\usepackage{booktabs}
\usepackage{amsmath}
\usepackage{subcaption}
\usepackage[margin=0.8in]{geometry}
\usepackage{tabularx}
\usepackage{algpseudocode}
\usepackage{amssymb}
\usepackage{multirow}
\usepackage{xcolor}
\usepackage{enumitem}
\modulolinenumbers[5]
\usepackage{changepage}
\usepackage{threeparttable}
\usepackage{microtype}
\usepackage{graphicx}
\usepackage{booktabs} 
\usepackage{hyperref}
\usepackage{algpseudocode} 
\usepackage{fancyhdr}
\usepackage{algorithm}
\setcitestyle{square}
\usepackage{amsfonts,bm}
\journal{Pattern Recognition}

\biboptions{authoryear}

\usepackage{setspace}

\begin{document}

\begin{frontmatter}


\title{Generating Relevant Counter-Examples from a Positive Unlabeled Dataset for Image Classification}

\author[affiliation3]{Florent Chiaroni$^{a,}$\corref{firstcorr}}
\cortext[firstcorr]{Corresponding author: Florent Chiaroni (permanent address: f.chiaroni@net.estia.fr) }
\author[affiliation2]{Ghazaleh Khodabandelou}
\author[affiliation1]{Mohamed-Cherif Rahal}
\author[affiliation4]{Nicolas Hueber}
\author[affiliation3]{Frederic Dufaux}
\address[affiliation1]{ VEDECOM Institute, Department of delegated driving (VEH08), Perception team, 
23 bis Allee des Marronniers, 78000 Versailles, France,  \\ \{florent.chiaroni, mohamed.rahal\}@vedecom.fr }
\address[affiliation2]{ VEDECOM Institute, Department of new uses (MOB05), Artificial Intelligence team, 23 bis Allee des Marronniers, 78000 Versailles, France, \\ ghazaleh.khodabandelou@vedecom.fr }
\address[affiliation3]{ L2S-CNRS-CentraleSupelec-Univ Paris-Sud-Univ Paris-Saclay, 3 rue Joliot Curie, 91190 Gif-sur-Yvette, France, \\ \{florent.chiaroni, frederic.dufaux\}@l2s.centralesupelec.fr }
\address[affiliation4]{ French-German Research Institute of Saint-Louis (ISL), ELSI team, 5 Rue du General Cassagnou, 68300 Saint-Louis, \\ nicolas.hueber@isl.eu }
%
\begin{abstract}
With surge of available but unlabeled data, Positive Unlabeled (PU) learning is becoming a thriving challenge. This work deals with this demanding task for which recent GAN-based PU approaches have demonstrated promising results. Generative adversarial Networks (GANs) are not hampered by deterministic bias or need for specific dimensionality. However, existing GAN-based PU approaches also present some drawbacks such as sensitive dependence to prior knowledge, a cumbersome architecture or first-stage overfitting. To settle these issues, we propose to incorporate a biased PU risk within the standard GAN discriminator loss function. In this manner, the discriminator is constrained to request the generator to converge towards the unlabeled samples distribution while diverging from the positive samples distribution. This enables the proposed model, referred to as D-GAN, to exclusively learn the counter-examples distribution without prior knowledge. Experiments demonstrate that our approach outperforms state-of-the-art PU methods without prior by overcoming their issues.
\end{abstract}


\end{frontmatter} 


\section{Introduction}

Nowadays, the number of available labeled datasets dedicated to perception applications has considerably augmented \citep{ILSVRC15}, \citep{DBLP:journals/corr/YuZSSX15}, \citep{Cordts2016Cityscapes}. However, when learning methods trained on these datasets are applied on real data, their performances are likely to deteriorate. Consequently, it is necessary to use a dataset specialized for the given target application. It turns out that it can be easy to get unlabeled data in some applications domains such as autonomous driving. Positive Unlabeled (PU) learning, also called \textit{partially supervised classification} \citep{liu_partially_2002}, enables to use these unlabeled data in combination with labeled samples of our class of interest: the positive class. The interest is that unlabeled data can contain relevant counter-examples, also called negative examples\footnote{We use the term \textit{example} to design a single instance (i.e. item, observation) included in a sample set of data following a given distribution.}. The difficulty is that unlabeled data can also contain a fraction $\pi_p$ of unlabeled positive examples. \cite{sansone_efficient_2018} enumerates several learning problems which can be addressed in this way such as the challenging information retrieval task.

Several PU learning methods exist, some of them adapted to image classification. They are generally classified into two categories. The former is censoring PU learning, formalized by \cite{elkan_learning_2008} and recently improved by \cite{northcutt_learning_2017}. The latter is case-control PU learning, introduced by \cite{ward_presence_only_2009}, formalized by \cite{du_plessis_analysis_2014}, and then consecutively improved by \cite{du_plessis_convex_2015} and \cite{kiryo_positive-unlabeled_2017-1} to reduce the training computational cost and alleviate the overfitting issue. In the context of the proposed approach, we focus our attention in this article on the recently presented GAN-based PU approaches. Thus we classify PU learning approaches into the two following groups suggested by \cite{kiryo_positive-unlabeled_2017-1}: one-stage and two-stage PU methods. 

One-stage PU methods such as the unbiased PU method (uPU) \citep{du_plessis_convex_2015} and the non-negative PU method (nnPU) \citep{kiryo_positive-unlabeled_2017-1} consist in training a classifier using an unbiased risk directly on the PU dataset. These methods have the advantage to need only one training of the classifier. However, they require dataset prior knowledge and consequently uPU and nnPU need to be combined with an approach estimating the prior knowledge \citep{jain_estimating_2016}, \citep{ramaswamy_mixture_2016}, \citep{christoffel_class_prior_2016}. Consequently, they are critically sensitive to slight prior variations per minibatch, as shown experimentally in Section \ref{subsubsec:Prior_noise_insensitivity}. 

Two-stage PU methods prepare during the first stage a Positive Negative (PN) dataset. For example, Rank Pruning method (RP) \citep{northcutt_learning_2017} firstly estimates the prior such that it can select
only the examples considered as the most confident, in order to substitute the unlabeled samples for the second-stage training of the classifier. RP achieves two-stage state-of-the-art performances without prior knowledge. However, it can only exploit a sub part of the training PU dataset. This can curb its prediction performances on complex datasets like CIFAR-10.
%
Recently, a new subcategory of two-stage PU methods appeared: GAN-based PU methods. They address the PU learning challenge by producing, thanks to an adversarial training \citep{goodfellow_generative_2014}, generated samples from a PU dataset during the first step. Then, they are used to train a standard Positive Negative (PN) classifier during the second step. 

We discuss in more details the above introduced PU methods uPU \citep{du_plessis_convex_2015}, nnPU \citep{kiryo_positive-unlabeled_2017-1}, RP \citep{northcutt_learning_2017}, GenPU \citep{hou_generative_2018_1} and PGAN \citep{chiaroni_learning_2018} in the related work Section \ref{sec:Related_work}.

We can nonetheless already make the following remarks, motivating the design of the proposed approach. Unbiased methods \citep{kiryo_positive-unlabeled_2017-1}, \citep{du_plessis_convex_2015}, and GenPU \citep{hou_generative_2018_1} are by definition sensitive to the prior knowledge in order to deal with a PU dataset. Conversely, whereas the two-stage censoring methods, such as RP \citep{northcutt_learning_2017},  do not require prior information, they suffer from generalization and unstability problems due to their selective process.
The PGAN method is the first that does not need prior knowledge nor a selective process, thus preserving a low sensitivity to prior knowledge combined with a training stability.
However, as mentioned in the PGAN article, it inherently suffers from first-stage overfitting. 
Based on these considerations, we propose in this article a novel GAN-based model, referred to as Divergent-GAN (D-GAN), to overcome the latter issue while preserving the PGAN advantages. To the best of our knowledge, we are the first to propose a GAN-based method to capture exclusively the unlabeled negative samples distribution from a PU dataset without prior knowledge. More specifically, our contributions are the 
following:
\begin{itemize}
	\item We propose to incorporate a biased PU learning loss function inside the original GAN \citep{goodfellow_generative_2014} discriminator loss function. The intuition behind it is to have the generative model solving the PU learning problem formulated in the discriminator loss function. In this way, the generator learns the distribution of the examples which are both unlabeled and not positive, namely the negative ones included in the unlabeled dataset;
	\item In addition, we study normalization techniques compatibility with the proposed framework. A learning model which manipulates different minibatches distributions should not use batch normalization techniques \citep{ioffe_batch_2015}. Alternative normalization techniques are discussed and experimented.
\end{itemize}
%
Consequently,  the proposed D-GAN framework compares favorably with PU learning state-of-the-art performances on simple MNIST \citep{lecun_gradient_based_1998} and complex CIFAR-10 \citep{krizhevsky_learning_2009} image datasets. The proposed framework code is available \footnote{D-GAN code for RGB images of $64 \times 64$ pixels is provided in supplementary material for reviewers}.

The remaining of this paper is structured as follow. Section \ref{sec:Related_work} presents previous PU learning approaches. Section \ref{sec:Method} describes the proposed method. Section \ref{sec:Experiments} presents the corresponding experimental results. Finally, in Section \ref{sec:Conclusion}, we draw conclusions and discuss perspectives.

\section{Related work}
\label{sec:Related_work}

The PU learning problem consists in trying to distinguish positive samples from negative samples by using a PU dataset. Let $X \in \mathbb{R}^m $ be the input random variable and $Y \in \{0, 1\}$ its associated label. $X$ can be a positive $X_P$, negative $X_N$ or unlabeled $X_U$ sample which respectively follow the distributions $p_P = p(X|Y=0)$, $p_N = p(X|Y=1)$ and $p_U = (1-\pi_P) \cdot p_N + \pi_P \cdot p_P$. The unknown prior $\pi_P \in (0, 1)$ represents the fraction of unlabeled positive examples included in the unlabeled dataset.

Previous works on PU learning \citep{denis_pac_1998} consider the entire distribution of the unlabeled examples as negative. In this way, all the negative examples, present in the unlabeled dataset, are always considered as negative. However, concerning the positive examples, it implies associating two contradictory labels to the distribution of positive examples in unknown proportions depending on the $\pi_P$ value. Thus, training directly a classifier with positive and unlabeled data provokes a bias in the training estimator, which is not present during a standard positive negative training. This bias can limit prediction performances of the learning model.

Several strategies have been proposed to solve this drawback such as unbiased methods \citep{du_plessis_analysis_2014}, \citep{du_plessis_convex_2015}, \citep{kiryo_positive-unlabeled_2017-1}, pruning method \citep{northcutt_learning_2017}, and more recently GAN based methods \citep{hou_generative_2018_1}, \citep{chiaroni_learning_2018}. However, those strategies still present some issues including prior knowledge sensitivity, training unsteadiness, or overfitting problems.

We present in this section different state-of-the-art methods and their respective drawbacks that we aim at overcoming with the proposed GAN-based PU framework.

\subsection{Unbiased methods}

In order to palliate a biased training, the authors of unbiased techniques \citep{du_plessis_analysis_2014}, \citep{du_plessis_convex_2015}, \citep{kiryo_positive-unlabeled_2017-1} suggest to avoid the estimator bias by adding some terms in the training loss function.
Then, the classifier behaves as if it is trained with a positive negative dataset. The authors firstly used a non convex loss function \citep{du_plessis_analysis_2014}, which then has been adapted for convex loss functions \citep{du_plessis_convex_2015} in order to reduce the computational burden. Subsequently, it was proposed to overcome the training overfitting by adding a binary condition (an "if" condition) in the training loss function \citep{kiryo_positive-unlabeled_2017-1}.

\begin{figure}[htb!]
\begin{center}
\begin{minipage}[c]{0.99\linewidth}
  \centering
  \centerline{\resizebox{8.75cm}{!}{\includegraphics[width=\linewidth]{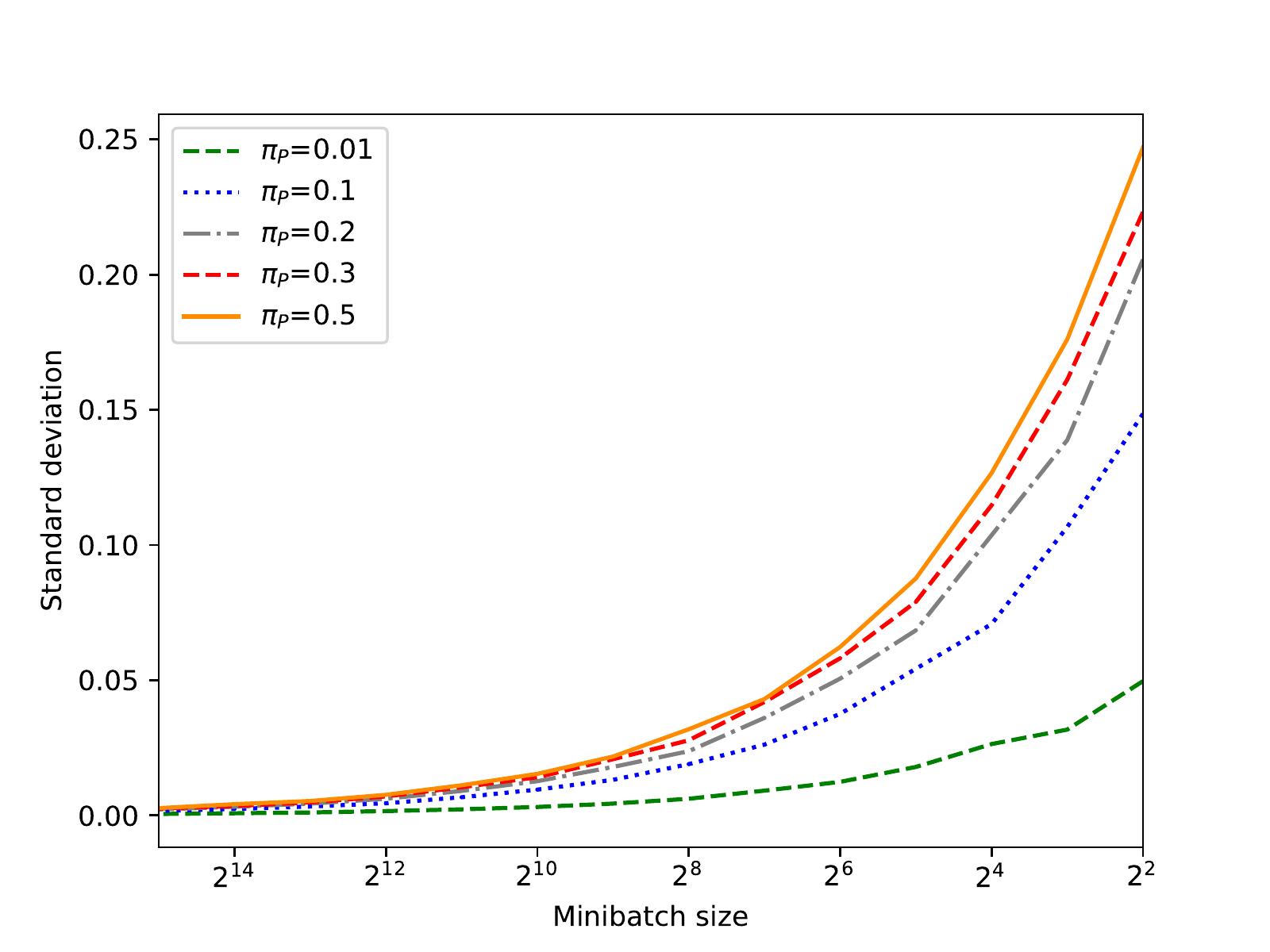}}}
\end{minipage}
\caption{Standard deviation per minibatch of the global prior $\pi_P$  in function of the minibatch size, for a given uniformly mixed dataset composed by 60000 examples.}
\label{fig:std_prior_in_function_of_batch_size}
\end{center}
%
\end{figure}

These methods exploit the prior $\pi_P$ in the empirical training loss function. However, we observe that the empirical prior value $\hat{\pi}_P$ per batch of small size (minibatch) is slightly different to $\pi_P$, as its standard deviation depends on the minibatch size, such that:
\begin{equation}
\hat{\pi}_P = \pi_P + \alpha,
\end{equation}
with $\alpha \sim p_{\alpha}(m)$, where $p_{\alpha}$ is the probability distribution of the noise $\alpha$ depending on the minibatch size $m$, as shown in Figure \ref{fig:std_prior_in_function_of_batch_size}. We observe that the worst case scenario is when $\pi_P$ is close to the value $0.5$, combined with a small batch size. The cases where $\pi_P$ is higher than $0.5$ behave symmetrically to the cases where $\pi_P$ is smaller than $0.5$.

In our case, we want to train a deep learning model using the stochastic gradient descent (SGD) optimization technique, which is known to be relevant with batches of small size. So the theoretical formulation of unbiased techniques cannot be maintained using SGD with small batch sizes. We will show empirically that in practice, unbiased techniques are highly sensitive to the minibatch size in terms of prediction performances, as they are theoretically sensitive to the prior $\pi_P$.

It turns out that it is possible to avoid this limitation with two-stage approaches.

\subsection{Two-stage approaches}

Two-stage approaches mainly consist in preparing during the first-stage a positive negative (PN) training dataset which then will be used to directly train a standard classifier during the second stage. One interest of those approaches is that they are not sensitive to the prior knowledge variation. Consequently, they are compatible with the use of minibatches, and thus are suitable when applying SGD optimization.

\subsubsection{Pruning approach}
Rank Pruning (RP) method \citep{northcutt_learning_2017} is a two-stage technique. It first estimates the prior $\pi_P$ and exploits it to prune the dataset in order to capture only a subset corresponding to the most confident positive and negative samples. Then, during the second stage it considers this subset as a cleanly labeled positive negative dataset to train a classifier. While not requiring prior knowledge in input, RP achieved state-of-the-art results for information retrieval in the One-vs-Rest task on simple datasets such as MNIST. However, by using a pruning strategy, RP can miss some relevant training examples not included in the selected subset of training. As a consequence, this can limit its generalization, as will be shown experimentally on Table \ref{tab:meanF1Score}, where RP is shown to be relatively unstable when compared to GAN-based approaches in terms of prediction performances. Using only a training subset is also a weakness on complex datasets like CIFAR-10, where a large training dataset is preferable to obtain better results.

Some approaches have been more recently proposed by exploiting generative adversarial networks (GANs) benefits, maintaining or increasing the prediction scores over the same PU learning tasks.

\subsubsection{GAN-based approaches}

GAN-based PU approaches represent a recent subcategory of two-stage PU methods, as proposed in GenPU \citep{hou_generative_2018_1} and PGAN \citep{chiaroni_learning_2018}. The interest of using GANs is twofold. First, GANs enable relevant data augmentation, as will be experimentally demonstrated on Table \ref{tab:meanF1Score}. Second, it allows for the use of high-level feature metrics to evaluate generated samples quality, thanks to the adversarial training. This can ease to capture a target distribution in a meaningful manner.
%

In this PU learning context, the generated samples replace the unlabeled ones by learning on the latter as PGAN \citep{chiaroni_learning_2018}, or on both unlabeled and positive labeled ones as GenPU \citep{hou_generative_2018_1}. Both methods exploit GANs benefits, but the functioning are different and they are not suitable under the same datasets conditions:

\begin{itemize}
    \item GenPU \citep{hou_generative_2018_1} is based on the original GAN convergence \citep{goodfellow_generative_2014}, such that: $\pi_P \cdot p_{G_P} + (1-\pi_P) \cdot p_{G_N} \xrightarrow[]{} p_U$, with $p_{G_P}$ the distribution of positive samples generated by the generator $G_P$, $p_{G_N}$ the distribution of the negative samples generated by the generator $G_N$, and $p_U$ the distribution of real unlabeled samples. 
    In practice, GenPU is an interesting PU method on simple datasets with few positive labeled samples, and it generates relevant counter-examples. However, training adversarially five learning models instead of two as in the original GAN framework \citep{goodfellow_generative_2014} to address \textit{standard} PU learning challenge\footnote{We use the term \textit{standard} to refer to the case where we have enough positive labeled examples (at least 100), such that the difficulty is mainly the ability to exploit counter-examples included in the unlabeled set.} is more computational demanding and not necessary to generate relevant counter-examples. Moreover, using five models amplifies the mode collapse issue, and the corresponding training optimization functions need three additional hyper-parameters combined with prior knowledge. This is impractical in the context of real applications where hyper-parameters tuning may be required on limited computational resources to adapt the model for a given application dataset.

    \item PGAN \citep{chiaroni_learning_2018} is trained to converge towards the unlabeled dataset distribution during the first step. During the second step, it exploits GANs imperfections for capturing the unlabeled distribution, such that the generated distribution at the adversarial equilibrium is still separable from the unlabeled samples distribution by a classifier. It presents a relatively steadier behaviour and better prediction performances than the two-stage baseline RP method on the complex RGB image dataset CIFAR-10 without prior knowledge. However, it is less suitable for relatively simpler datasets like MNIST. The problem is that the generated samples are all considered as negative samples by the classifier. But this is possible only if the generated samples distribution converges close enough towards the unlabeled samples distribution, while not matching it. If the PGAN first-stage performs as expected theoretically by \citep{goodfellow_generative_2014}, then the PGAN classification second stage falls back into the initial PU learning problem.
\end{itemize}

Our proposed approach, presented in Sec. \ref{sec:Method}, overcomes previously enumerated PU methods shortcomings, to address the \textit{standard} PU learning task, as summarized on Table \ref{tab:contributions}.

\begin{table*}[h]
\centering
\resizebox{16.75cm}{!}{
\begin{tabular}{c|ccccc}
\toprule
Methods & D-GAN (proposed) & PGAN \citep{chiaroni_learning_2018} & GenPU \citep{hou_generative_2018_1} & RP \citep{northcutt_learning_2017} & nnPU \citep{kiryo_positive-unlabeled_2017-1} \\
\midrule
No need of priori knowledge & $\surd$ & $\surd$ & & $\surd$ & \\
\midrule
No first-stage overfitting & $\surd$ & & $\surd$ & $\surd$ & $\surd$ \\
\midrule
Generalizable over complex datasets & $\surd$ & $\surd$ & & & \\
\midrule
Able to generate relevant counter-examples & $\surd$ & & $\surd$ & & \\
\midrule
Training stability using SGD & $\surd$ & $\surd$ & & $\surd$ & \\
\midrule
Original GAN architecture & $\surd$ & $\surd$ & & & \\
\midrule
Code availability & $\surd$ & & & $\surd$ & $\surd$ \\
\bottomrule
\end{tabular}}
\caption{Summary of presented state-of-the-art methods advantages and drawbacks compared to the proposed D-GAN approach. A void cell means that the mentioned criterium is not applicable with the corresponding method.}
\label{tab:contributions}
\end{table*}

\section{Proposed Approach}
\label{sec:Method}

In this section, we first briefly recall the main reasoning which motivated our research work. Next, we discuss some features of a biased PU risk. We then propose to incorporate this risk into a generic GAN framework in order to guide the generator convergence towards the negative samples distribution, denoted as $p_N$, included inside the unlabeled dataset distribution, denoted as $p_U$. Furthermore, we study regularization techniques to manipulate three distinct types of minibatches: positive, unlabeled and generated ones.
%
%
%

\subsection{Motivation}

In PU learning, if a classifier associates a given expected label value with positive examples, and in parallel associates a second distinct label value with unlabeled examples, then it is proven that the negative non-labeled examples are exclusively associated with the label of non-labeled examples \citep{denis_pac_1998}, \citep{blum_combining_1998}, \citep{lee_learning_2003}. 
Concerning GANs, it has been shown that the discriminator learning task influences directly the adversarial generator behaviour \citep{mao_least_2017}. 
%
%
%
%
%
%

Based on these considerations, this work aims at incorporating a biased PU risk inside the traditional GAN discriminator cost function. This compels the discriminator $D$, to separate negative from positive distributions, which in turn guides the generator $G$, to exclusively learn the unlabeled counter-examples distribution from a PU dataset.
As a matter of the fact, the proposed method is novel in the way it exclusively generates relevant counter-examples without prior knowledge information, while preserving a standard GAN architecture.
%
%
%

Thereafter, we present the biased PU risk that we incorporate in the proposed GAN PU discriminator training loss function.

\subsection{Biased PU risk to incorporate}
\label{subsec:Biased_PU_risk}


In what follows, we first explain the expected PU functionality to be incorporated into the GAN discriminator loss function.
\textbf{Biased PU risk setting:} Let $D:{\mathbb{R}^m \to [0,1]} $ be the decision function which is, later on, considered as the discriminator $D$, of the proposed framework network. We have $l(\hat{y}, y)$ such that $l: {[0,1] \times [0, 1] \to \mathbb{R}}$ is the arbitrary cost function with the predicted output $\hat{y}$ of $D$ for a given example and the corresponding label $y$ as input. $D$ is trained with a PU risk $R_{PU}$ to predict the label value $1$ for the unlabeled examples, and the label value $0$ for the positive labeled ones such that:
\begin{equation}
\begin{aligned}
R_{PU}(D) =  \mathbb{E}_{x_U \sim \bm{p_U}} [l(D(x_U), {1})] 
          + \mathbb{E}_{x_P \sim  {p_P}} [l(D(x_P), {0})].
\end{aligned}
\label{eq:PU_risk}
\end{equation}
Given the composition of the distribution $p_U$, we develop:
\begin{equation}
\begin{aligned}
R_{PU}(D) = (1-\pi_P) \cdot \mathbb{E}_{x_N \sim  {p_N}} [l(D(x_N), {1})] 
          + \pi_P \cdot \mathbb{E}_{x_P \sim  {p_P}} [l(D(x_P), {1})]  
           + \mathbb{E}_{x_P \sim  {p_P}} [l(D(x_P), {0})].
\end{aligned}
\label{eq:PU_risk_developped}
\end{equation}

\textbf{Counter-examples are correctly labeled:} Decomposed in this way, the negative examples included in the unlabeled dataset are associated exclusively to the label value $1$ for any $\pi_P$ value, such that the negative training examples are all correctly labeled.

When there is no overfitting on training positive examples, then one can assume that labeled and unlabeled positive examples follow the same distribution $p_P$, as mentioned in \citep{kiryo_positive-unlabeled_2017-1}. Since expectations are linear, $p_P$ is associated to both contradictory labels $0$ and $1$ as below:
\begin{equation}
\begin{aligned}
R_{PU}(D)=  \mathbb{E}_{x_N \sim  {p_N}} [(1-\pi_P) l(D(x_N), {1})] 
+ \mathbb{E}_{x_P \sim  {p_P}} [\pi_P l(D(x_P), {1}) + l(D(x_P), {0})].
\end{aligned}
\label{eq:PU_risk_developed_positive}
\end{equation}

\textbf{Positive samples distribution $p_P$ is shifted away from the counter-examples distribution $p_N$:} When defining the cost-function $l$ as the binary cross-entropy $H$ (Eq. \ref{eq:Binary_cross_entropy}) such that $l=H$, then we can demonstrate that the second term in the Equation \ref{eq:PU_risk_developed_positive} is equivalent to associating the positive distribution $p_P$ with a unified  biased intermediate label value $\delta$. The binary cross-entropy $H$ is defined as:
\begin{equation}
\begin{aligned}
H(D(X),Y) = - Y log(D(X)) - (1-Y) log(1-D(X)),
\end{aligned}
\label{eq:Binary_cross_entropy}
\end{equation}
where $Y$ is the label value associated with the input $X$ of $D$. If $l=H$, then concerning the second term of the Equation \ref{eq:PU_risk_developed_positive}, we can demonstrate that:
\begin{equation}
\begin{aligned}
\pi_P H(D(x_P),1) + 1 H(D(x_P),0) =& -\pi_P log(D(x_P)) - 1 log(1-D(x_P)) \\
=& -\pi_P log(D(x_P)) - (1+\pi_P-\pi_P)log(1-D(x_P)) \\
%
%
=& (1+\pi_P) \cdot \bigg[-\frac{\pi_P}{1+\pi_P}log(D(x_P))  -(1-\frac{\pi_P}{1+\pi_P})log(1-D(x_P))\bigg] \\
=& (1+\pi_P) \cdot H \big( D(x_P),\frac{\pi_P}{1+\pi_P} \big) \\
=& (1+\pi_P) \cdot H(D(x_P), {\delta}),
\end{aligned}
\label{eq:Demonstration_PU_risk_developed_positive_with_delta}
\end{equation}
with $\delta=\pi_P/(1+\pi_P)$. Consequently, the PU risk becomes:
\begin{equation}
\begin{aligned}
R_{PU}(D) = \mathbb{E}_{x_N \sim  {p_N}} [(1-\pi_P) H(D(x_N), {1})] 
          + \mathbb{E}_{x_P \sim  {p_P}} [(1+\pi_P) H(D(x_P), {\delta})].
\end{aligned}
\label{eq:PU_risk_developed_positive_with_delta}
\end{equation}
Such a PU risk has been previously called biased or constrained in the literature \citep{du_plessis_analysis_2014}, \citep{liu_partially_2002}. The equivalence between Equations \ref{eq:PU_risk_developed_positive} and \ref{eq:PU_risk_developed_positive_with_delta} makes it possible to estimate the restricted interval of possible values for $\delta$ without using prior such that if $\pi_P \in (0,1)$ then:
\begin{equation}
\begin{aligned}
0<\pi_P<1 \Leftrightarrow 0<\delta<\frac{1}{1+1}.
\end{aligned}
\label{eq:Boundary_estimation}
\end{equation}
In other words, $\delta \in (0,1/2)$. This confirms that for any $\pi_P$ value between $0$ and $1$, labeled and unlabeled positive examples are associated with a label value $\delta$ comprised between $0$ and $1/2$. Therefore, when training $D$ with the risk $R_{PU}$, the $D$ prediction related to the unlabeled positive examples is shifted away from the label value 1. From $D$ prediction output point of view, this risk makes the positive distribution $p_P$ \textit{diverging} from the negative distribution $p_N$. Thus, $D$ is trained to predict the label value $1$ exclusively for the counter-examples.

\subsection{Proposed generative model}

The insight in the proposed D-GAN model can be expressed as follows: $D$ addresses to $G$ the riddle: \textit{Show me what IS unlabeled AND NOT positive.} It turns out that negative examples included in the unlabeled dataset are both unlabeled and not positive. Consequently, $G$ addresses this riddle by learning to show the negative samples distribution to $D$.

\textbf{GAN background:} We first give a short recall of the original GAN discriminator. It is trained to distinguish real unlabeled samples distribution $p_U$ from generated samples distribution $p_G$ with the loss function $L_{D_{GAN}}$ defined as:
\begin{equation}
\begin{aligned}
    L_{D_{GAN}}(G,D) = \mathbb{E}_{x_U \sim p_U} [-log⁡D(x_U)]
    + \mathbb{E}_{z \sim p_z} [-log⁡(1-D(G(z)))],
\end{aligned}
\label{eq:Original_GAN_D_loss_function}
\end{equation}
where $z$ stands for the input random vector of the generative model $G$ such that $G(z)$ is a generated sample. $z$ follows a uniform or normal distribution.
It turns out that the binary cross-entropy formulation (Eq. \ref{eq:Binary_cross_entropy}) implies $H(D(X),1)=-log(D(X))$ and $H(D(X),0)=-log(1-D(X))$. Consequently, $L_{D_{GAN}}$ can be expressed as follows:
\begin{equation}
\begin{aligned}
    L_{D_{GAN}}(G,D) =& \mathbb{E}_{x_U \sim p_U} [H(D(x_U),1)] 
    + \mathbb{E}_{z \sim p_z} [H(D(G(z)),0)].
\end{aligned}
\end{equation}

\textbf{Towards a GAN biased discriminator loss function:} The proposed approach aims at training $G$ to learn the negative samples distribution $p_N$ instead of learning the distribution $p_U$. This replaces the standard GAN task \textit{\textquotedblleft Show me what is unlabeled \textquotedblright}, by the task \textit{\textquotedblleft Show me what is both unlabeled and not positive\textquotedblright}. We now propose to incorporate the benefits of a biased PU risk (Eq. \ref{eq:PU_risk}) into the original GAN discriminator loss function (Eq. \ref{eq:Original_GAN_D_loss_function}). To this end, we define the D-GAN discriminator loss function $L_D$ by adding the term $\mathbb{E}_{x_P \sim p_P} [ H(D(x_P),0) ]$ to $L_{D_{GAN}}$. Consequently, in the proposed D-GAN framework, the training discriminator loss function $L_D$ of $D$ becomes:
\begin{equation}
\begin{aligned}
    L_D(G,D) =& L_{D_{GAN}}(G,D) + \mathbb{E}_{x_P \sim p_P} [ H(D(x_P),0) ].
\end{aligned}
\end{equation}
If we develop the term $L_{D_{GAN}}$, we then obtain:  
\begin{equation}
\begin{aligned}
    L_D(G,D) =& \mathbb{E}_{x_U \sim p_U} [H(⁡D(x_U),1)]
    + \mathbb{E}_{z \sim p_z} [H(D(G(z)),0)] 
    + \mathbb{E}_{x_P \sim p_P} [ H(D(x_P),0) ] \\
    =& R_{PU}(D) + \mathbb{E}_{z \sim p_z} [ H(D(G(z)),0) ].
\end{aligned}
\end{equation}
In other words, the $R_{PU}$ risk (Eq. \ref{eq:PU_risk}) is incorporated inside the D-GAN discriminator loss function. To this extent, $D$ can be trained to only consider the counter-examples as the most real examples by associating to them exclusively the label value $1$. This can be considered as applying a constrained optimization.

\textbf{The generator generates the counter-examples distribution:} In contrast, the role of $G$ during the adversarial training is to generate samples considered by $D$ as $1$. As suggested by \citep{goodfellow_generative_2014}, the training loss function $L_G$ of $G$ is such that:
\begin{equation}
\begin{aligned}
L_G (G,D) &= \mathbb{E}_{z \sim p_z} [ -log(D(G(z))) ] \\
&= \mathbb{E}_{z \sim p_z} [ H(D(G(z)),1) ].
\end{aligned}
\end{equation}
As developed previously, we recall that $D$ exclusively considers the negative examples as $1$ thanks to the $R_{PU}$ risk presented previously. Thus, if $D$ trainable weights are fixed in the proposed framework, then we propose to reinterpret in $L_G$ the label value $1$ as $D(x_N)$, as follows:
\begin{equation}
\begin{aligned}
L_G (G,D) &= \mathbb{E}_{z \sim p_z, x_N \sim p_N} [ H(D(G(z)),D(x_N)) ] \\
&= \mathbb{E}_{z \sim p_z, x_N \sim p_N} [ -D(x_N)log(D(G(z))) ],
\end{aligned}
\end{equation}
such that the distance between the generated samples distribution and $p_N$ is minimized.
Consequently, this justifies the convergence of $G$ in the proposed D-GAN framework towards the negative samples distribution $p_N$, for any $\pi_P \in (0,1)$.

\textbf{Implementation:} The corresponding implementation algorithm \ref{alg:D_GAN_implementation} of the proposed first-stage D-GAN approach enables to adversarially train $D$ and $G$ to respectively minimize loss functions $L_D$ and $L_G$.

%
\begin{algorithm*}[tb]
   \caption{Minibatch SGD training of the D-GAN}
   \label{alg:D_GAN_implementation}
\begin{algorithmic}
   \State GAN training ($1^{st}$ step)
   
   \For{number of training iterations}  
   \State Sample minibatch of $m$ noise samples $\{ z^{(1)},..., z^{(m)} \}$ from noise prior $p_z$.
   \State Sample minibatch of $m$ unlabeled examples $\{ x_U^{(1)},..., x_U^{(m)} \}$ from data distribution $p_U$.
   \State Sample minibatch of $m$ positive labeled examples $\{ x_P^{(1)},..., x_P^{(m)} \}$ from data distribution $p_P$.
   \State Update $D$ by descending its stochastic gradient:
   \State $\begin{aligned} \nabla_{\theta_{D}}  \frac{1}{m} \sum_{i=0}^m \Big[ -log⁡D(x_U^{(i)})-log\big[1-D(G(z^{(i)}))\big] -log\big[1-D(x_p^{(i)})\big] \Big] \end{aligned}$
   \State Sample minibatch of $m$ noise samples $\{ z^{(1)},..., z^{(m)} \}$ from noise prior $p_z$.
   \State Update $G$ by descending its stochastic gradient:
   \State $\begin{aligned} \nabla_{\theta_{G}}  \frac{1}{m}\sum_{i=0}^m -log\big[D(G(z^{(i)}))\big] \end{aligned}$
   \EndFor
   \State Classifier training ($2^{nd}$ step):
   
   \For{number of training iterations}
   
   \State Sample minibatch of $m$ positive labeled examples $\{ x_P^{(1)},..., x_P^{(m)} \}$ from data distribution $p_P$.
   \State Sample minibatch of $m$ noise samples $\{ z^{(1)},..., z^{(m)} \}$ from data distribution $p_z$.
   \State Update $C$ by descending its stochastic gradient:
   \State $\begin{aligned} \nabla_{\theta_{C}}  \frac{1}{2 \cdot m} \sum_{i=1}^{m} \Big[ l(C(x_P^{(i)}),1) + l(C(G(z^{(i)})),0) \Big] 
   \end{aligned}$
   \EndFor
   \State The gradient-based updates can use any standard gradient-based learning rule. We use Adam in our experiments.
\end{algorithmic}
\end{algorithm*}

\begin{figure*}[h]
\centering
\centerline{\resizebox{15.75cm}{!}{\includegraphics[scale=0.35]{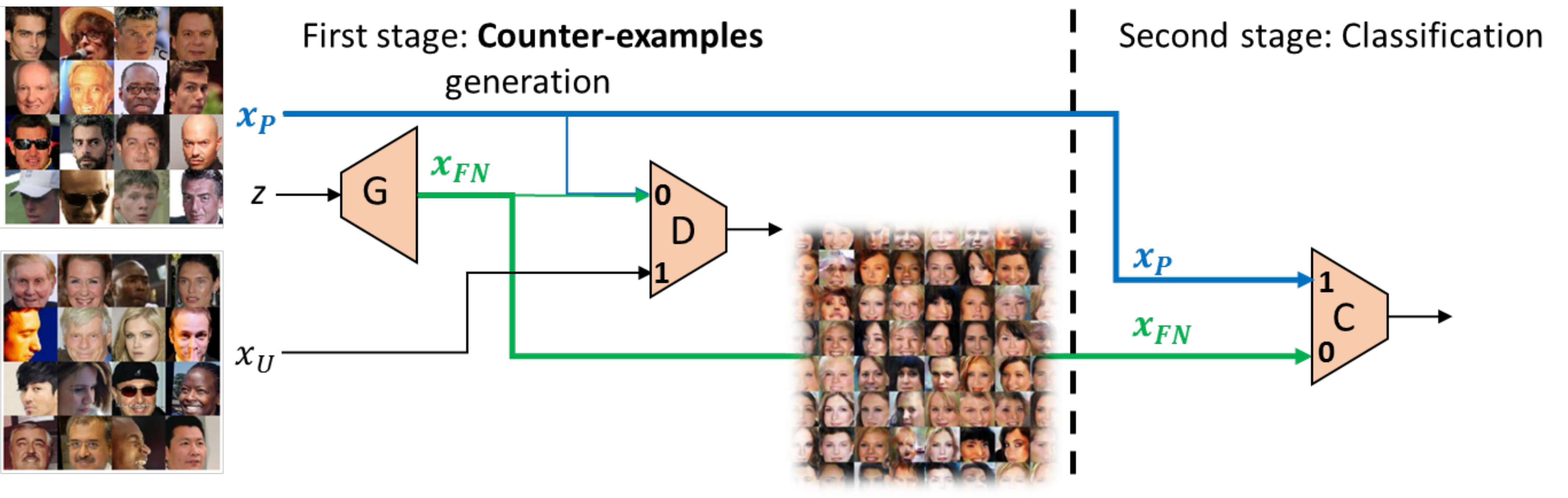}}}
\caption{Proposed GAN-based PU approach, $x_{FN}$ represents the generated samples which are similar to real negative samples $x_N$, $G$ is the generative model, $D$ is the discriminator, $C$ is the classifier used to perform the binary Positive-Negative (PN) classification.}
\label{fig:DIVERGENT-GAN_Architecture}
\end{figure*}

%
\textbf{Second-stage: Positive-Generative learning.} Once the D-GAN training is completed, the second step can be carried out. It consists in training a classifier $C$ to distinguish fake generated examples $x_{FN}=G(z)$, which are ideally equivalent to the real negative samples, from real positive labeled samples as illustrated in Figure \ref{fig:DIVERGENT-GAN_Architecture}.

In practice, the worst-case scenario is when $D$ overfits the positive examples during the adversarial training. Another pitfall is when $D$ cannot encode the complexity of the boundary between positive and negative examples included in the unlabeled dataset. In such cases, $D$ will consider some unlabeled positive examples as negative ones. As a consequence, this implies that $G$ will also generate some examples following a subset of the positive samples distribution. Thus, the D-GAN will tend to behave as the PGAN \citep{chiaroni_learning_2018}, which seems to be the best solution in this situation.

The next section presents effective regularization techniques to overcome these issues in the context of the proposed GAN-based PU framework.

\subsection{Discriminator regularizations}
\label{subsec:Discriminator_regularizations}

Nowadays, Batch Normalization (BN) \citep{ioffe_batch_2015} is considered as a one of the most relevant regularization techniques commonly used in deep neural networks architectures. Its utility for GANs training has been highlighted by \citep{radford_unsupervised_2015} for the DCGAN architecture in order to stabilize the adversarial training. Other variants like the Wasserstein-GAN \citep{arjovsky_wasserstein_2017-1} or the Loss-Sensitive GAN \citep{qi_loss_sensitive_2017} confirmed its interest. As developed in \citep{ioffe_batch_2015}, BN addresses issues like vanishing or exploding gradient problems, as well as the risk of getting stuck in a poor local minima, by reducing the internal covariate shift problem of the learning model. A higher learning rate can be used and it can significantly improve the training speed.

\textbf{Multiple minibatch manipulation incompatibility.} BN regularizes the model, in such as way that a training example (i.e. single instance) from a given minibatch sample is considered in conjunction with other examples of this minibatch sample. This is the consequence of estimating the mean and variance normalization parameters one time per minibatch, and then applying them on each example in the minibatch. When positive examples $x_P$ and unlabeled examples $x_U$ are not in the same training minibatch, as this is the case in our discriminator loss function, this does not enable to link labeled positive examples with the unlabeled positive ones. Consequently, this cannot produce a distance between positive and negative examples predictions. To counter this problem, we could imagine to apply BN on a unified minibatch which contains a fraction of each distribution $x_P$, $x_U$ and $x_F$. But the BN effect is greatly influenced by the content of the minibatch on which it is applied. Therefore, the fraction $\pi_P$ of positive examples included in $x_U$ will negatively impact the BN outcome. 
%

\textbf{Compatible normalization techniques:} However, BN benefits in a more traditional training are not negligible. Hence, we propose to use two alternative techniques in order to replace the BN role in the proposed GAN-based PU framework. On the one hand, \textbf{Layer Normalization} (LN) \citep{ba_layer_2016} is a frequently used technique with sequential networks, as it can be applied for each sequential example independently. With LN, the normalization for a given example is computed on its resulting output feature map layers, and the mean and variance are computed independently for each example of a minibatch. On the other hand, \textbf{Spectral Normalization} (SN) \citep{miyato_spectral_2018} is a recent competing technique for GANs \citep{miyato_spectral_2018} training which can stabilize the training of $D$ against input perturbations \citep{farnia_generalizable_2018} by perfoming a weight normalization. In this way, a training manipulating multiple types of minibatch distributions preserves SN effectiveness. For these reasons, we propose to apply LN or SN instead of BN inside our discriminative model structure. The use of these normalization techniques will be validated in Sec. \ref{sec:Experiments}.


\textbf{Dropout alleviates the positive overfitting problem:} As mentionned in the previous section, we can only deduce Equation \ref{eq:PU_risk_developed_positive} if we consider that the positive samples distribution is the same for both labeled and unlabeled ones. In practice, this assumption holds in the case of a large dataset, such that this overfitting problem concerning the positive examples disappears. The dropout \citep{srivastava_dropout_2014}, \citep{mordido_dropout_gan_2018} generalization technique is also a solution. In the context of the proposed D-GAN training, we introduce dropout in the top fully connected layer of $D$. We enable it during $D$ training steps, and conversely disable it during $G$ training steps. This improves the evaluation of generated samples which is transmitted from $D$ to $G$ by back-propagation. In the next section, we will show that dropout alleviates the positive examples overfitting during long D-GAN trainings. This insures to exclusively generate counter-examples.

The next section presents experimental results demonstrating the usefulness of the proposed approach.


%
%
%
%
%
%

\section{Experimental Results}
\label{sec:Experiments}




%
%

In this section, we assess the performance of the proposed approach. We first experimentally validate the expected discriminator prediction behaviour when it is applied on a positive unlabeled dataset (Sec. \ref{subsubsec:experimental_risk_analysis}), and study the impact of regularization (Sec. \ref{subsubsec:Without_BN_tests}). Then, we show the ability of the generator to generate counter-examples for different types of PU datasets, including two-dimensional points and natural RGB images (Sec. \ref{subsubsec:counter_examples_generation}). Finally, we evaluate the proposed model prediction robustness and compare it with state-of-the-art PU learning methods in terms of prior noise (Sec. \ref{subsubsec:Prior_noise_insensitivity}) and first-stage overfitting (Sec. \ref{subsubsec:One_vs_Rest_sssec}).

\subsection{Settings}

\begin{figure*}[h]
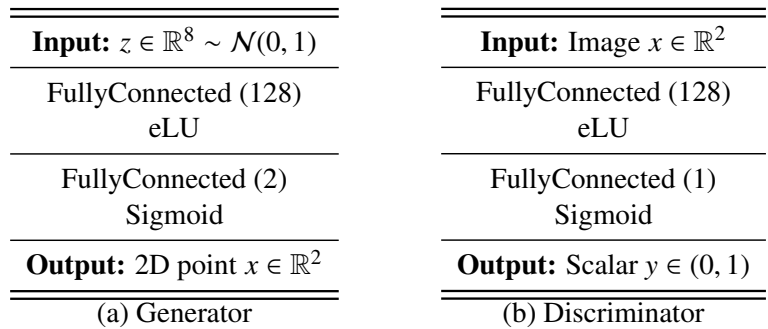

\begin{center}
\begin{minipage}[c]{0.32\linewidth}
  \centering
  \centering{
  	\begin{tabular}{c} 
    	\toprule
    	\toprule
    	\textbf{Input:} $z \in \mathbb{R}^{8} \sim \mathcal{N}(0,1)$ \\
    	\midrule
    	FullyConnected (128) \\
    	eLU \\
    	\midrule
    	FullyConnected (2) \\
    	Sigmoid \\
    	\midrule
    	\textbf{Output:} 2D point $x \in \mathbb{R}^{2}$ \\
    	\bottomrule
    	\bottomrule
  	\end{tabular}}
  \centerline{(a) Generator}\medskip
\end{minipage}
\begin{minipage}[c]{0.32\linewidth}
  \centering
  \centering{
  	\begin{tabular}{c} 
    	\toprule
    	\toprule
    	\textbf{Input:} Image $x \in \mathbb{R}^{2}$ \\
    	\midrule
    	FullyConnected (128) \\
    	eLU \\
    	\midrule
    	FullyConnected (1) \\
    	Sigmoid \\
    	\midrule
    	\textbf{Output:} Scalar $y \in (0,1)$\\
    	\bottomrule
    	\bottomrule
  	\end{tabular}}
  \centerline{(b) Discriminator}\medskip
\end{minipage}
\caption{Fully connected GAN model architecture used for two dimensional points datasets. Minibatch size 64, optimizer Adam. We trained the model during 100 epochs on 2D point datasets.}
\label{fig:archi_2D_points}
\end{center}
%
\end{figure*}

\begin{figure}[!ht]
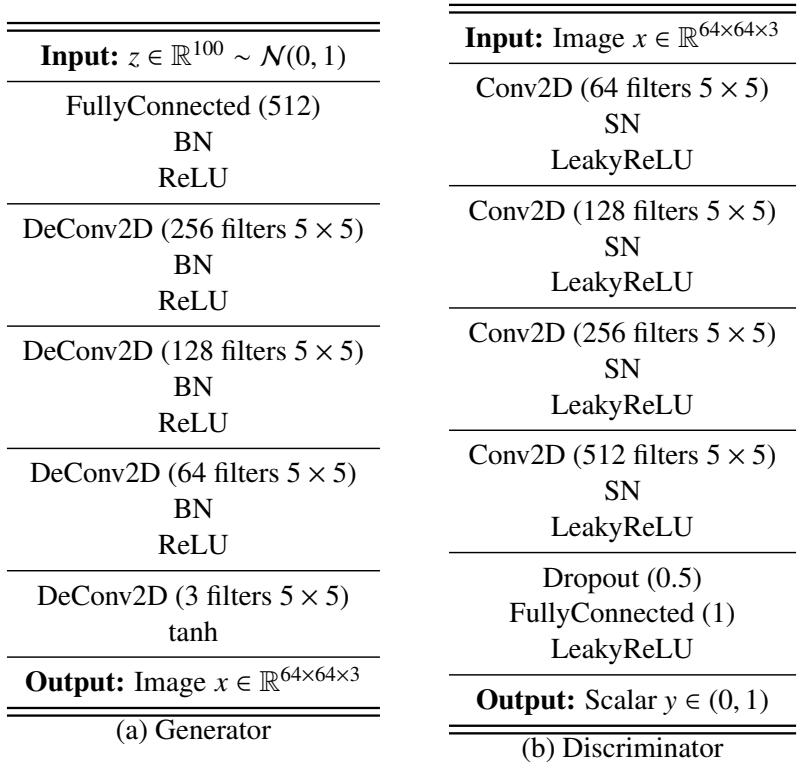

\begin{center}
\begin{minipage}[c]{0.32\linewidth}
  \centering
  \centering{
  	\begin{tabular}{c} 
    	\toprule
    	\toprule
    	\textbf{Input:} $z \in \mathbb{R}^{100} \sim \mathcal{N}(0,1)$ \\
    	\midrule
    	FullyConnected (512) \\
    	BN \\
    	ReLU \\
    	\midrule
    	DeConv2D (256 filters $5 \times 5$) \\
    	BN \\
    	ReLU \\
    	\midrule
    	DeConv2D (128 filters $5 \times 5$) \\
    	BN \\
    	ReLU \\
    	\midrule
    	DeConv2D (64 filters $5 \times 5$) \\
    	BN \\
    	ReLU \\
    	\midrule
    	DeConv2D (3 filters $5 \times 5$) \\
    	tanh \\
    	\midrule
    	\textbf{Output:} Image $x \in \mathbb{R}^{64 \times 64 \times 3}$ \\
    	\bottomrule
    	\bottomrule
  	\end{tabular}}
  \centerline{(a) Generator}\medskip
\end{minipage}
\begin{minipage}[c]{0.32\linewidth}
  \centering
  \centering{
  	\begin{tabular}{c} 
    	\toprule
    	\toprule
    	\textbf{Input:} Image $x \in \mathbb{R}^{64 \times 64 \times 3}$ \\
    	\midrule
    	Conv2D (64 filters $5 \times 5$) \\
    	SN \\
    	LeakyReLU \\
    	\midrule
    	Conv2D (128 filters $5 \times 5$) \\
    	SN \\
    	LeakyReLU \\
    	\midrule
    	Conv2D (256 filters $5 \times 5$) \\
    	SN \\
    	LeakyReLU \\
    	\midrule
    	Conv2D (512 filters $5 \times 5$) \\
    	SN \\
    	LeakyReLU \\
    	\midrule
    	Dropout (0.5) \\
    	FullyConnected (1) \\
    	LeakyReLU \\
    	\midrule
    	\textbf{Output:} Scalar $y \in (0,1)$\\
    	\bottomrule
    	\bottomrule
  	\end{tabular}}
  \centerline{(b) Discriminator}\medskip
\end{minipage}
\caption{Convolutional GAN model architecture used for 64*64 RGB images of celebA dataset. Minibatch size: 64, optimizer: Adam, 2D stride of $2 \times 2$. We trained the model during 100 epochs on the celebA dataset.}
\label{fig:archi_celebA}
\end{center}
%
\end{figure}

We detail in this section the settings of the experiments. We have adapted the first-stage discriminator and generator architectures of the proposed GAN based PU framework depending on the dataset on which they are applied, as follows:
\begin{itemize}
	\item \textbf{2D point dataset}: In order to deal with 2D point datasets, we use a GAN architecture composed of fully connected layers (FullyConnected). The generator and discriminator architectures are summarized in Figure \ref{fig:archi_2D_points}.
	\item \textbf{MNIST} \citep{lecun_gradient_based_1998}: In order to deal with grayscale images of dimension 28*28 pixels from the MNIST dataset, we use a deep convolutional GAN architecture (DCGAN) such that the generator contains transposed convolutional (DeConv2D) top layers, and the discriminator contains convolutional (Conv2D) bottom layers as illustrated in Figures \ref{fig:archi_MNIST_CIFAR_10} (a) and (b).
	\item \textbf{CIFAR-10} \citep{krizhevsky_learning_2009}: In order to deal with RGB images of size 32*32 pixels from the CIFAR-10 dataset, we use the same DCGAN architecture presented in Figures \ref{fig:archi_MNIST_CIFAR_10} (a) and (b). We only adapt the feature maps size depending on the width (w), the height (h), and the number of channels (ch) of input RGB images.
	\item \textbf{celebA} \citep{liu2015faceattributes}: In order to deal with RGB images of size 64*64 from the celebA dataset, we use a deeper convolutional GAN architecture presented in Figure \ref{fig:archi_celebA}.
\end{itemize}

\begin{figure*}[h]
\begin{center}
\begin{minipage}[c]{0.32\linewidth}
  \centering
  \centering{
  	\begin{tabular}{c} 
    	\toprule
    	\toprule
    	\textbf{Input:} $z \in \mathbb{R}^{100} \sim \mathcal{N}(0,1)$ \\
    	\midrule
    	FullyConnected (1024) \\
    	BN \\
    	ReLU \\
    	\midrule
    	FullyConnected ($128 \times \frac{h}{4} \times \frac{w}{4}$) \\
    	BN \\
    	ReLU \\
    	\midrule
    	DeConv2D (64 filters $4 \times 4$) \\
    	BN \\
    	ReLU \\
    	\midrule
    	DeConv2D (\textit{ch} filters $4 \times 4$) \\
    	Sigmoid \\
    	\midrule
    	\textbf{Output:} Image $x \in \mathbb{R}^{h \times w \times ch}$\\
    	\bottomrule
    	\bottomrule
  	\end{tabular}}
  \centerline{(a) Generator}\medskip
\end{minipage}
\begin{minipage}[c]{0.32\linewidth}
  \centering
  \centering{
  	\begin{tabular}{c} 
    	\toprule
    	\toprule
    	\textbf{Input:} Image $x \in \mathbb{R}^{h \times w \times ch}$ \\
    	\midrule
    	Conv2D (64 filters $4 \times 4$) \\
    	SN \\
    	LeakyReLU \\
    	\midrule
    	Conv2D (128 filters $4 \times 4$) \\
    	SN \\
    	LeakyReLU \\
    	\midrule
    	FullyConnected (1024) \\
    	SN \\
    	LeakyReLU \\
    	\midrule
    	Dropout (0.5) \\
    	FullyConnected (1) \\
    	Sigmoid \\
    	\midrule
    	\textbf{Output:} Scalar $y \in (0,1)$\\
    	\bottomrule
    	\bottomrule
  	\end{tabular}}
  \centerline{(b) Discriminator}\medskip
\end{minipage}
\begin{minipage}[c]{0.32\linewidth}
  \centering
  \centering{
  	\begin{tabular}{c} 
    	\toprule
    	\toprule
    	\textbf{Input:} Image $x \in \mathbb{R}^{h \times w \times ch}$ \\
    	\midrule
    	Conv2D (32 filters $5 \times 5$) \\
    	ReLU \\
    	Maxpooling ($2 \times 2$) \\
    	\midrule
    	Conv2D (64 filters $5 \times 5$) \\
    	ReLU \\
    	Maxpooling ($2 \times 2$) \\
    	\midrule
    	FullyConnected (1024) \\
    	ReLU \\
    	\midrule
    	Dropout (0.5) \\
    	FullyConnected (2) \\
    	Softmax \\
    	\midrule
    	\textbf{Output:} One hot vector $y \in (0,1)^2$\\
    	\bottomrule
    	\bottomrule
  	\end{tabular}}
  \centerline{(b) Classifier}\medskip
\end{minipage}
\caption{Convolutional GAN model architecture used for 28*28 grayscale MNIST and 32*32 RGB CIFAR-10 image datasets. For MNIST we set h=28, w=28, ch=1. For CIFAR-10 we set h=32, w=32, ch=3. Minibatch size: 64, optimizer: Adam, strides of $2 \times 2$ for the generator Deconv2D and the discriminator Conv2D layers, strides of $1 \times 1$ for the classifier Conv2D layers. We trained the model during 40 epochs and 1000 epochs respectively on MNIST and CIFAR-10 datasets.}
\label{fig:archi_MNIST_CIFAR_10}
\end{center}
%
\end{figure*}

Concerning the PU dataset initialization from a standard PN dataset, in all the experiments, except the ones in Sec. \ref{subsubsec:Prior_noise_insensitivity}, we use the methodology proposed by \citep{chiaroni_learning_2018}. More specifically, we set $\rho=0.5$ which is the fraction of positive labeled examples of the initial PN dataset that we unlabel such that they are included into the unlabeled dataset. Then, we set $\pi_P$ which is the fraction that represents these unlabeled positive examples among the unlabeled dataset. This method is interesting for testing an approach depending on $\pi_P$, independantly of the selected fraction $1-\rho$ of positive labeled samples.
%
%
%
%
%

\subsection{Qualitative analysis}
\label{subsec:qualitative_analysis}

We start by studying qualitatively whether the discriminator behaves as expected in practice. More precisely, we need to verify whether it exclusively associates the counter-examples distribution with the label value 1, and the positive samples distribution with an intermediate label value between $0$ and $1/2$. 

In Sec. \ref{subsubsec:experimental_risk_analysis}, we start by showing the relation between the PU loss function and the proposed equivalent PN loss function including a biased label for positive examples, as mentioned in Sec. \ref{subsec:Biased_PU_risk}.
Then, in Sec. \ref{subsubsec:Without_BN_tests}, we investigate which regularization techniques enable to preserve the same behaviour on an image dataset such that the discriminator does not suffer from overfitting during the epoch training iterations.

\subsubsection{Empirical Positive Unlabeled risk analysis}
\label{subsubsec:experimental_risk_analysis}

We have previously demonstrated (Eq. \ref{eq:Demonstration_PU_risk_developed_positive_with_delta}) that we can reformulate the discriminator PU training loss function $R_{PU}$ into a PN training loss function, referred to as $R_{PN}$, by replacing the two opposite labels $0$ and $1$ associated to positive samples distribution $p_P$ by an intermediate label value $\delta$ depending on $\pi_P$, such that we obtain:
\begin{equation}
R_{PU}(D) = R_{PN}(D),
\label{eq:PU_equal_biased_PN}
\end{equation}
with:
\begin{equation}
\begin{aligned}
\left\{
    \begin{array}{ll}
R_{PU}(D) =& \mathbb{E}_{x_U \sim \bm{p_U}} [H(D(x_U),\bm{1})]  
           + \mathbb{E}_{x_P \sim \bm{p_P}} [H(D(x_P),\bm{0})], \\
R_{PN}(D) =& \mathbb{E}_{x_N \sim \bm{p_N}} [(1-\pi_P) H(D(x_N),\bm{1})] 
           + \mathbb{E}_{x_P \sim \bm{p_P}} [(1+\pi_P) H(D(x_P),\bm{\delta})].
    \end{array}
\right.
\end{aligned}
\label{eq:PU_and_PN_risks_formula}
\end{equation}

%
%
%
%

\begin{figure}[!ht]
\begin{center}
\begin{minipage}[c]{0.49\linewidth}
  \centering
  \centerline{\includegraphics[width=\linewidth]{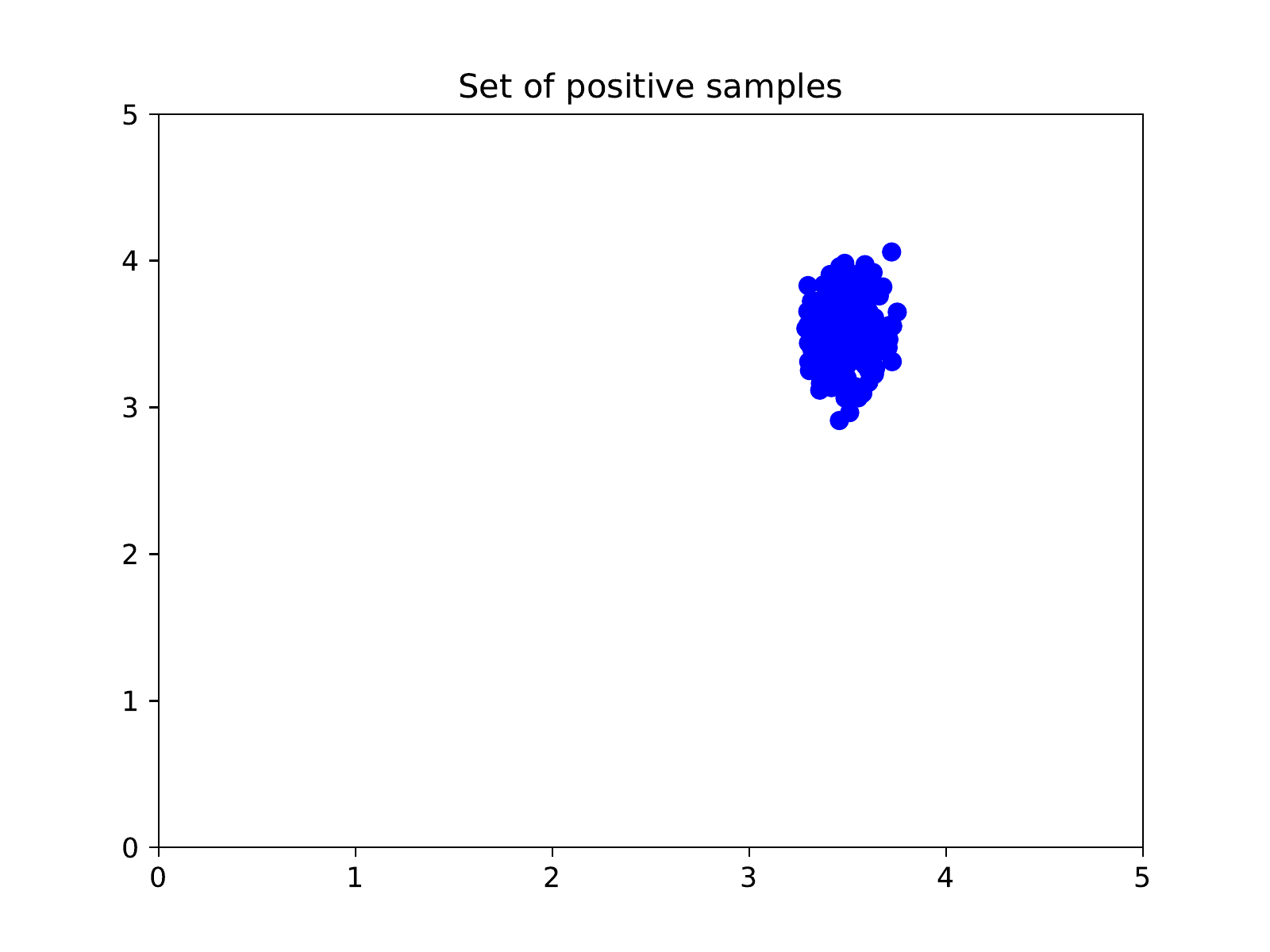}}
  \centerline{(a)}\medskip
\end{minipage}
\begin{minipage}[c]{0.49\linewidth}
  \centering
  \centerline{\includegraphics[width=\linewidth]{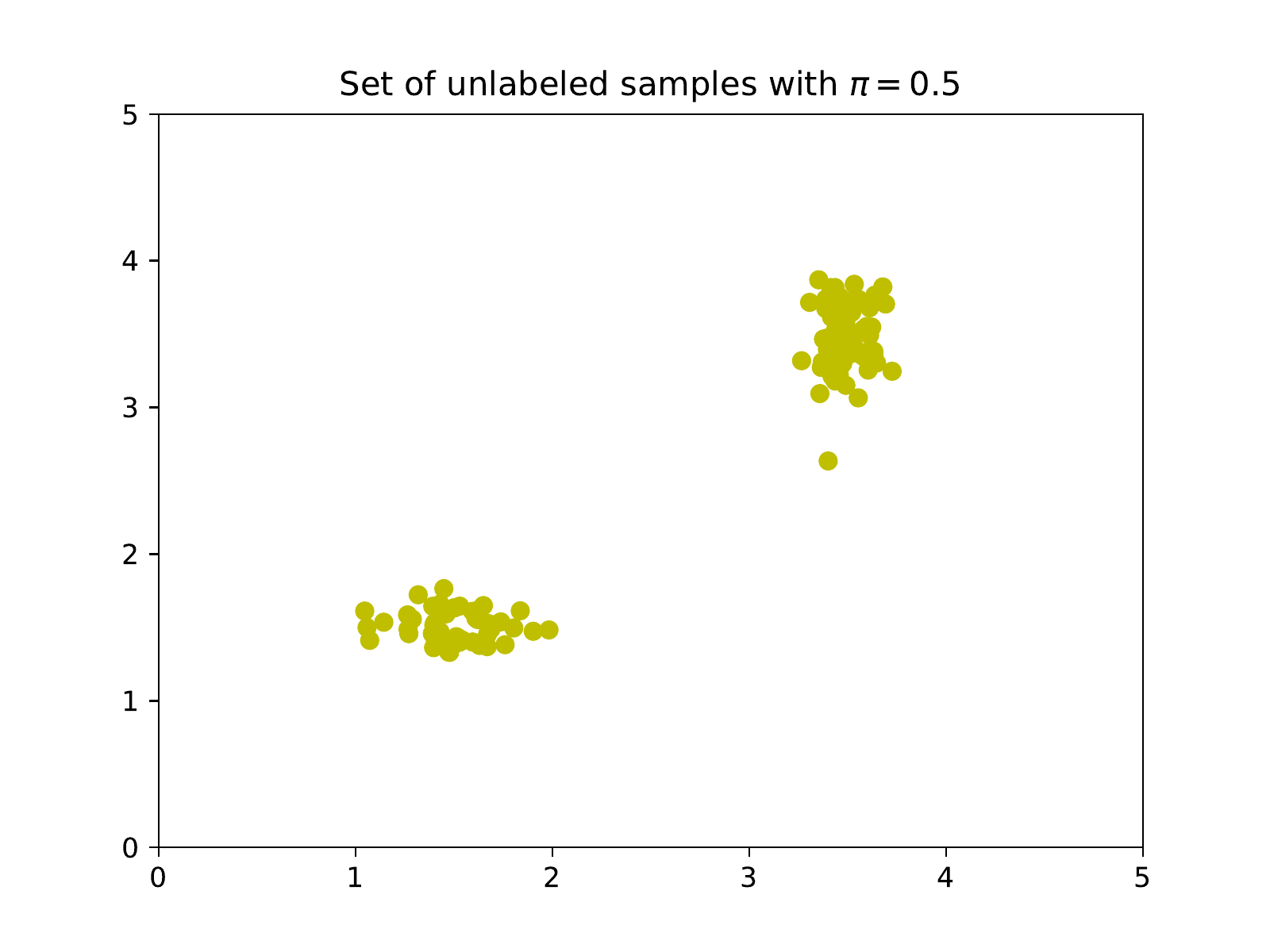}}
  \centerline{(b)}\medskip
\end{minipage}
\begin{minipage}[c]{0.49\linewidth}
  \centering
  \centerline{\includegraphics[width=\linewidth]{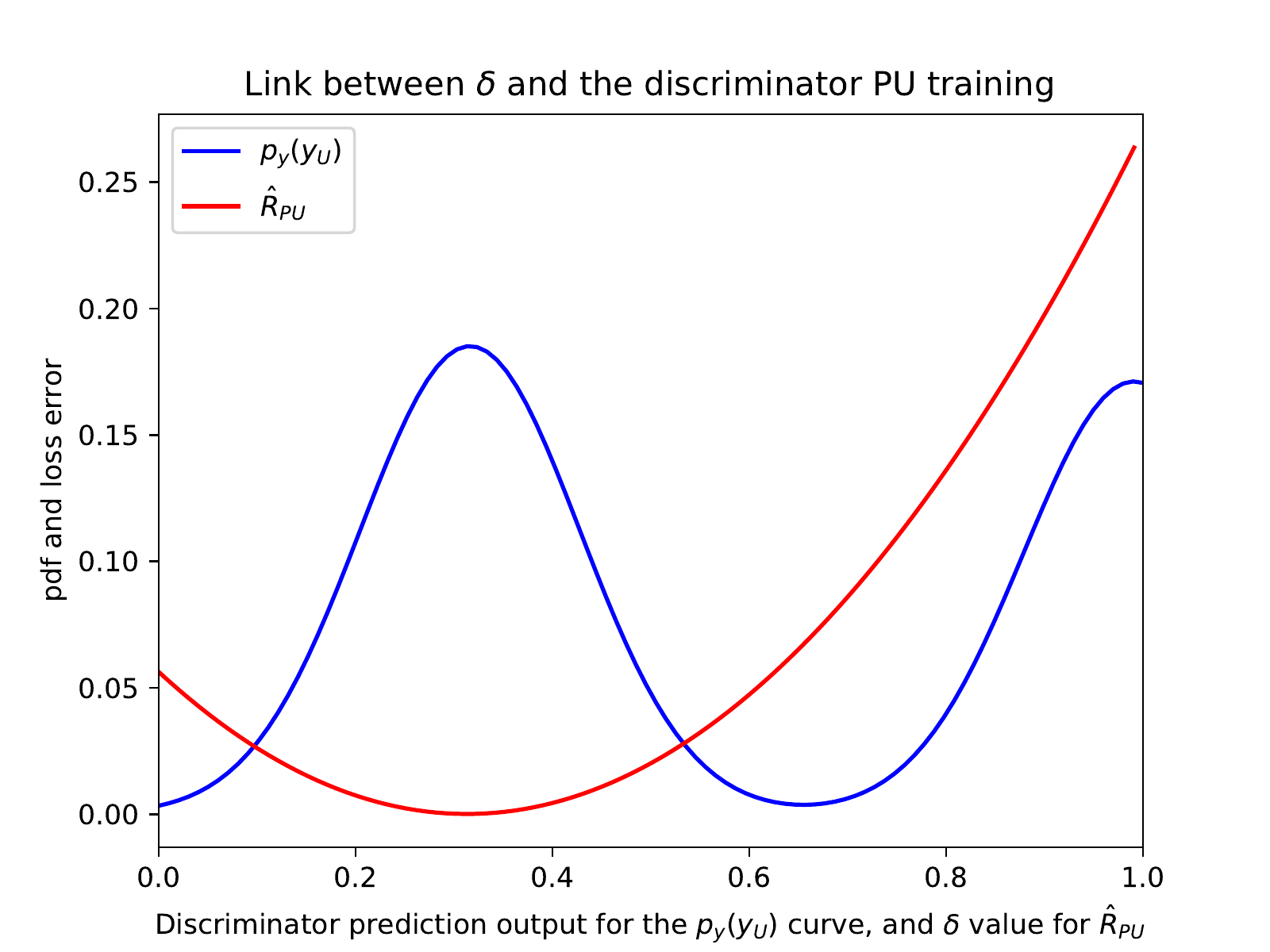}}
  \centerline{(c)}\medskip
\end{minipage}
\begin{minipage}[c]{0.49\linewidth}
  \centering
  \centerline{\includegraphics[width=\linewidth]{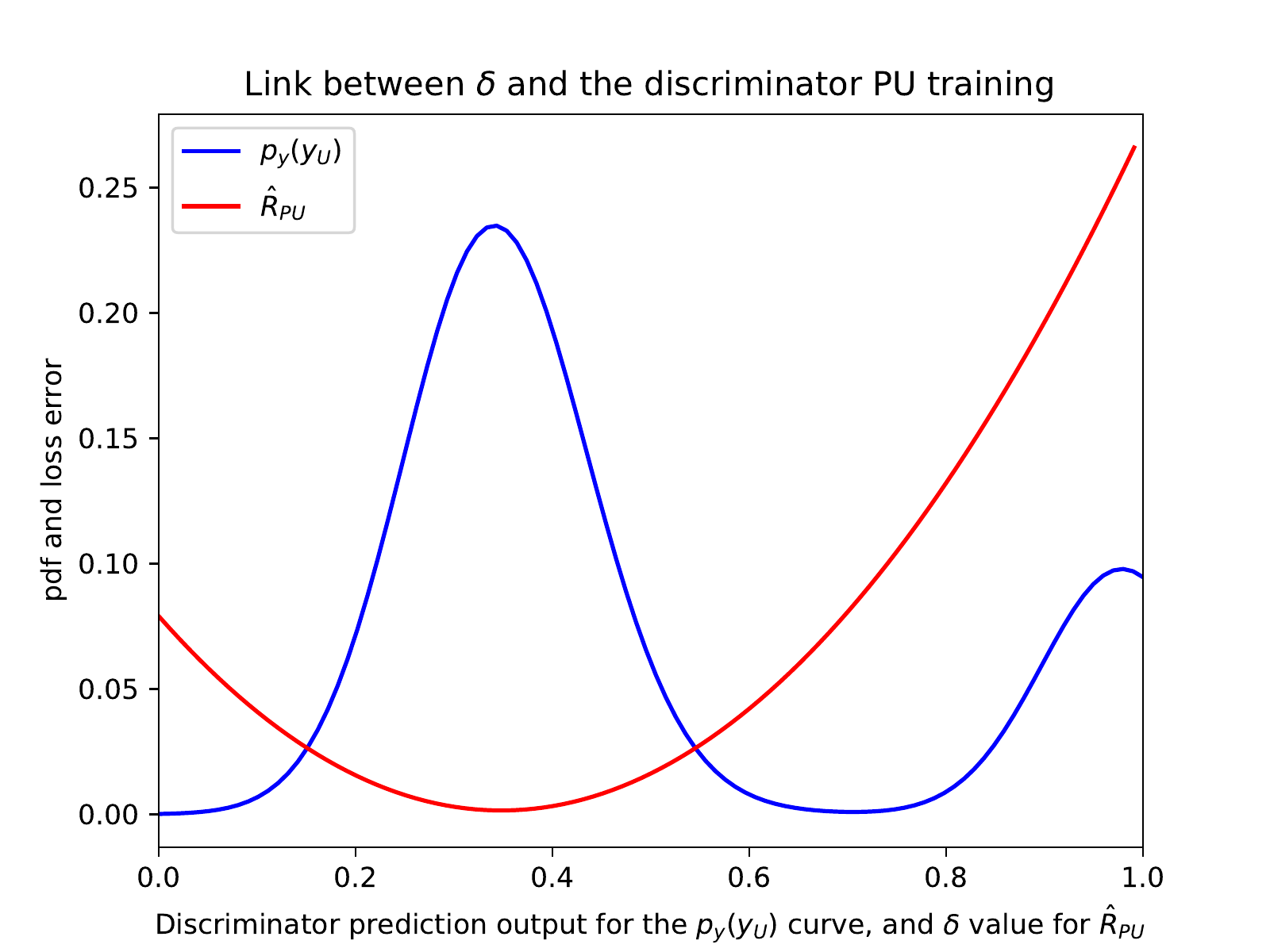}}
  \centerline{(d)}\medskip
\end{minipage}
\caption{Link between the PN loss function suggested (Eq. \ref{eq:PU_equal_biased_PN}) and the distribution of the discriminator output predictions for an input training minibatch. For this experiment,  
$D$ is a multi-layer perceptron. $D$ has been trained to distinguish a 2D gaussian distribution to another one by using the risk $R_{PU}$ on a PU dataset. (a) Shows a set of 2D points considered as positive samples. (b) Shows a set of 2D points considered as unlabeled samples. Both curves in (c) and (d) have been normalized to get a better visualization. For (c), $p_Y(y_U)$ (in blue), with $y_U=D(x_U)$, represents the probability distribution of $D$ predicted outputs for a minibatch of unlabeled samples, with $\pi_P=0.5$. $\hat{R}_{PU}(D)$ (in red) represents the PN risk computed in function of $\delta$ with the $R_{PN}$ proposed Equation \ref{eq:PU_and_PN_risks_formula} on a minibatch of positive and negative labeled samples, once $D$ is trained with $R_{PU}$ risk (Eq. \ref{eq:PU_risk}). (d) shows the same curves as in (c) but by giving in input a concatenation of an unlabeled minibatch with a positive labeled minibatch. Unlabeled positive and labeled positive samples provide a unified prediction output distribution.}
\label{fig:PU_risk_behaviour}
\end{center}
%
\end{figure}

It turns out that we can verify the same relation empirically. As illustrated in Figure \ref{fig:PU_risk_behaviour} with 2D point samples following gaussian distributions, if we train the discriminator $D$ with a multilayer perceptron structure using the PU loss function $R_{PU}$, then its predictions outputs for an unlabeled batch sample are partitioned in the vicinity of two different labels. Positive examples are centered around an intermediate label value corresponding to $\delta$. Conversely, $D$ output predictions for the negative examples are centered around the label value $1$. In addition, we have also computed the approximated PN risk $\hat{R}_{PN}$ using negative labeled and positive labeled samples, for several $\delta$ values between $0$ and $1$. We can observe that the global minimum of the PN approximated risk $\hat{R}_{PN}$ as a function of $\delta$ corresponds graphically to the global maximum of the density function corresponding to $D$ output predictions for a positive set. This coincides also with the equality presented in Equation \ref{eq:PU_equal_biased_PN}.

To sum up, this illustrates experimentally that if $D$ is trained with the $R_{PU}$ loss function, then it should predict the label value $1$ exclusively for the negative samples, which is the necessary condition to guide the generator during the adversarial training to learn exclusively the counter-examples distribution. 

However, this behaviour is only possible if $D$ does not overfit labeled and unlabeled positive samples. In other words, $D$ should be able to discriminate unlabeled positive examples from the unlabeled negative ones. Therefore, in order to generalize the proposed GAN framework to image datasets, we compare in the next section some state-of-the-art regularization techniques commonly used in deep learning models, in order to select the most appropriate one.
%
%

\subsubsection{Impact of regularizations on the discriminator}
\label{subsubsec:Without_BN_tests}

\begin{figure}[!ht]
\begin{center}
\begin{minipage}[c]{0.24\linewidth}
  \centering
  \centerline{\includegraphics[width=\linewidth]{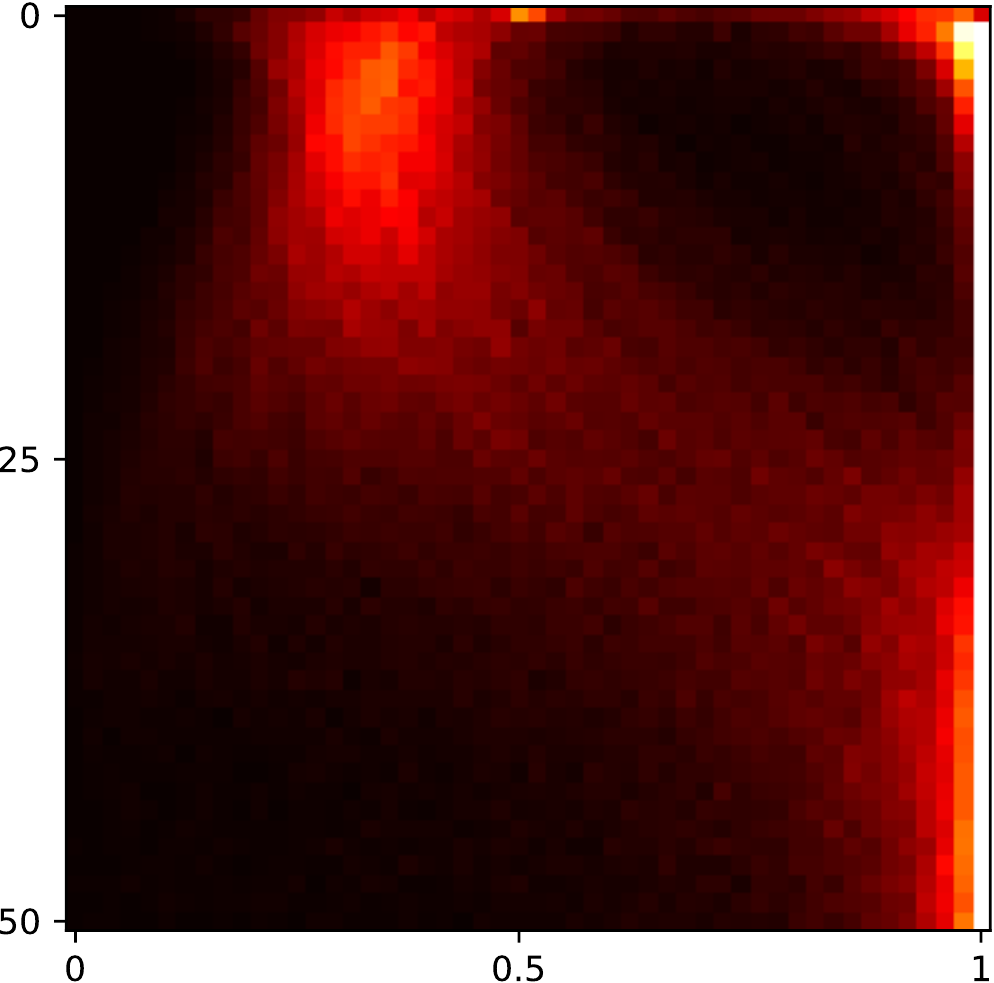}}
  \centerline{(a) No norm}\medskip
\end{minipage}
\begin{minipage}[c]{0.24\linewidth}
  \centering
  \centerline{\includegraphics[width=\linewidth]{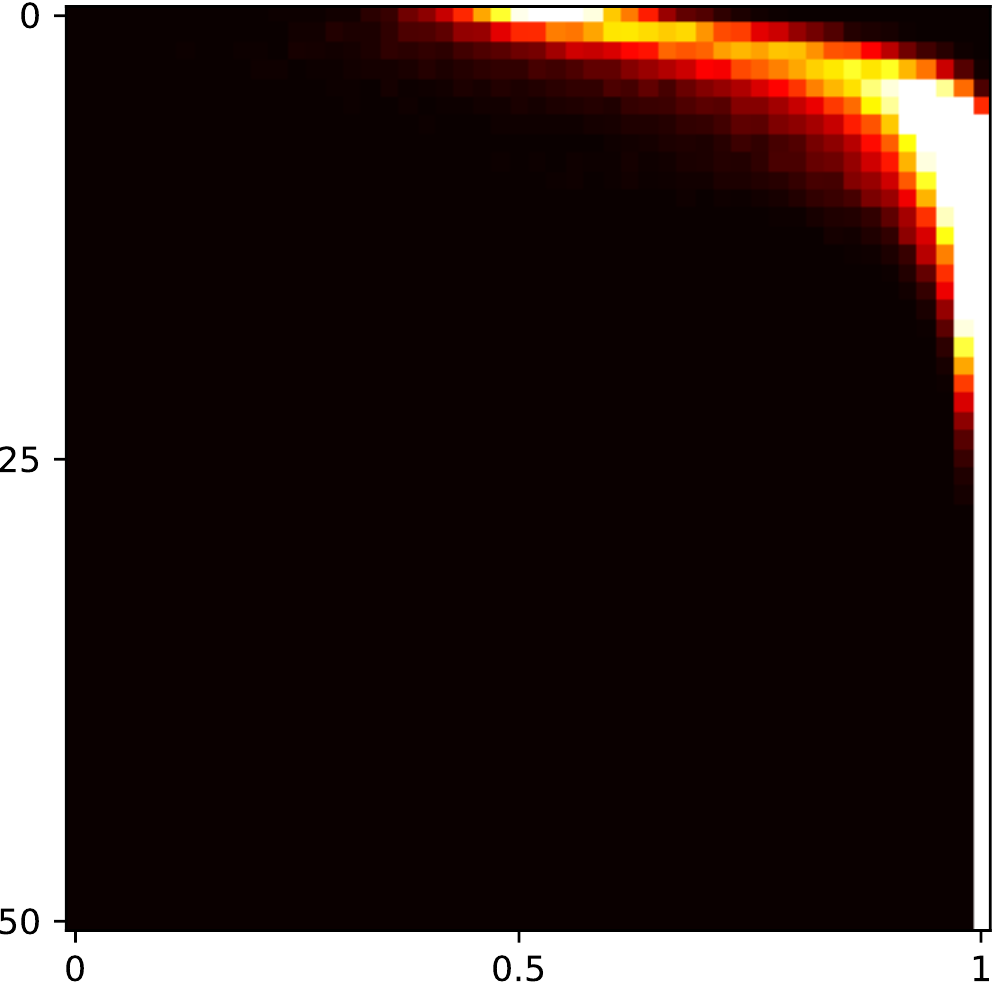}}
  \centerline{(b) BN}\medskip
\end{minipage}
\begin{minipage}[c]{0.24\linewidth}
  \centering
  \centerline{\includegraphics[width=\linewidth]{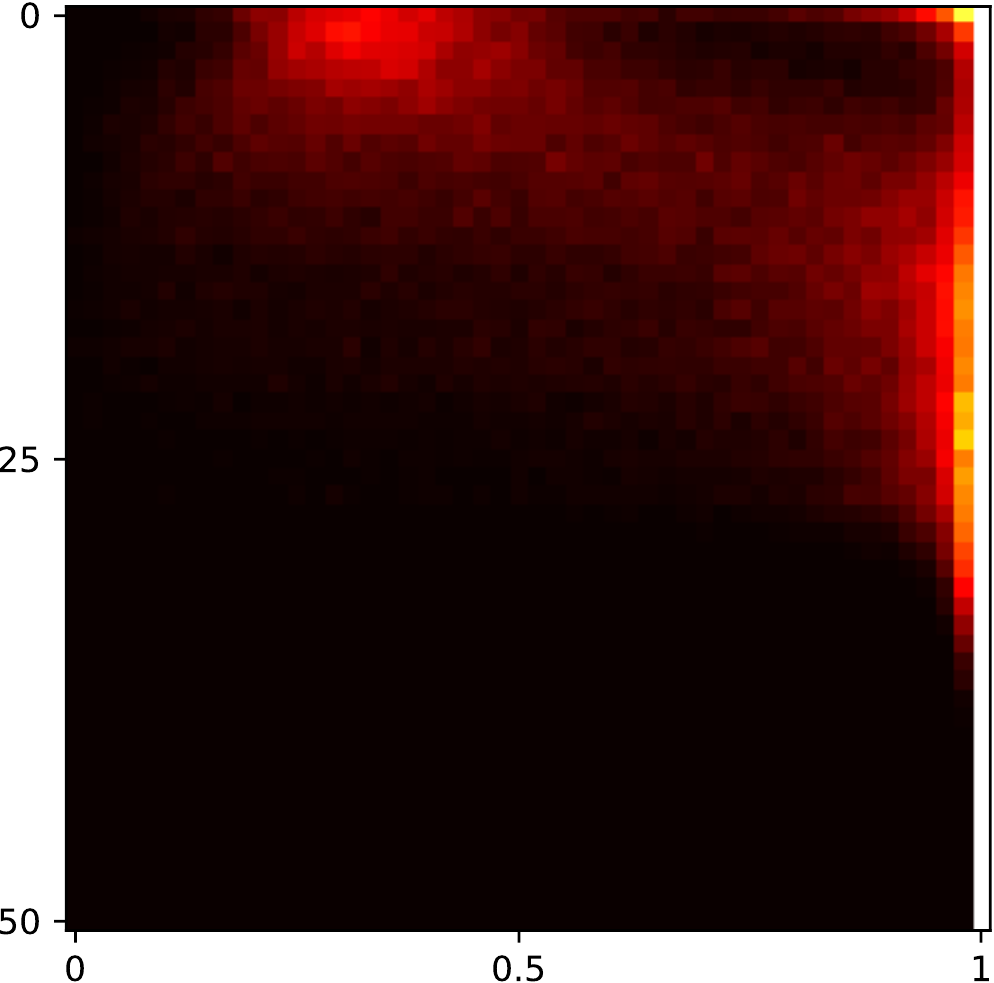}}
  \centerline{(c) LN}\medskip
\end{minipage}
\begin{minipage}[c]{0.24\linewidth}
  \centering
  \centerline{\includegraphics[width=\linewidth]{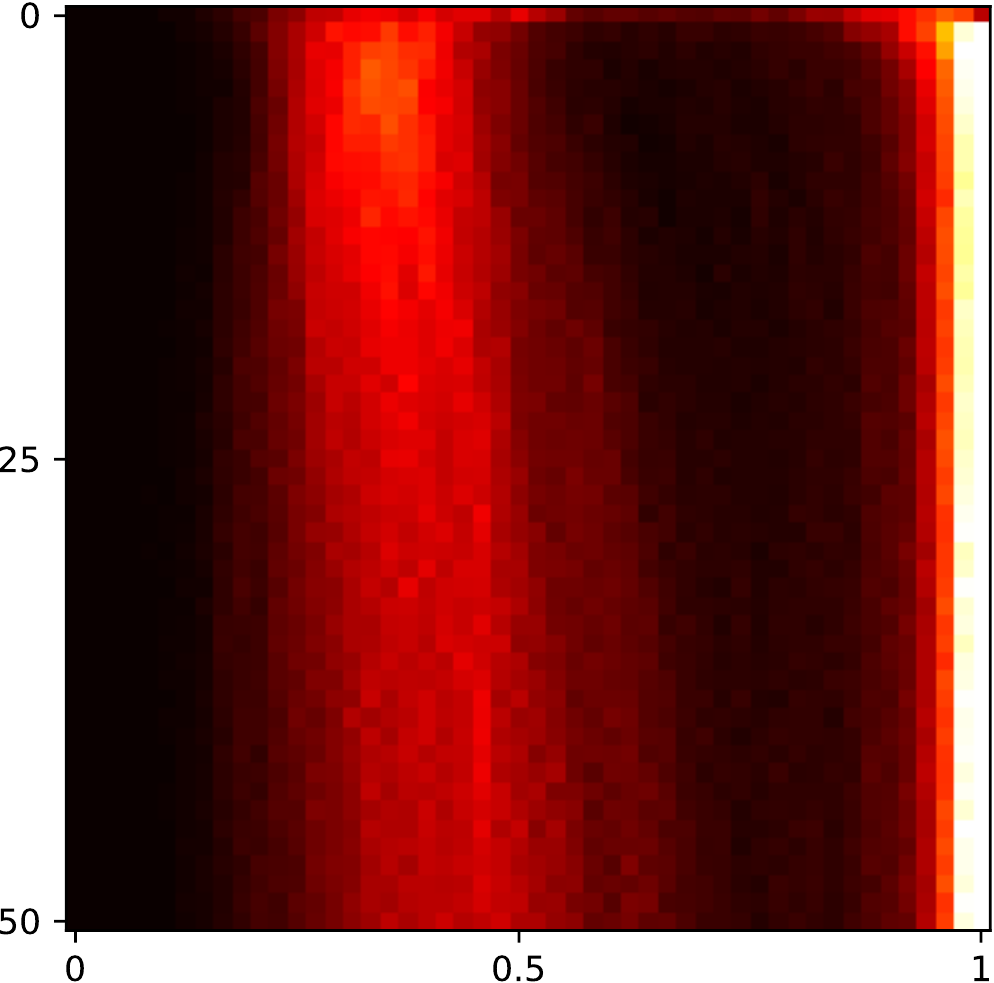}}
  \centerline{(d) SN}\medskip
\end{minipage}
\begin{minipage}[c]{0.24\linewidth}
  \centering
  \centerline{\includegraphics[width=\linewidth]{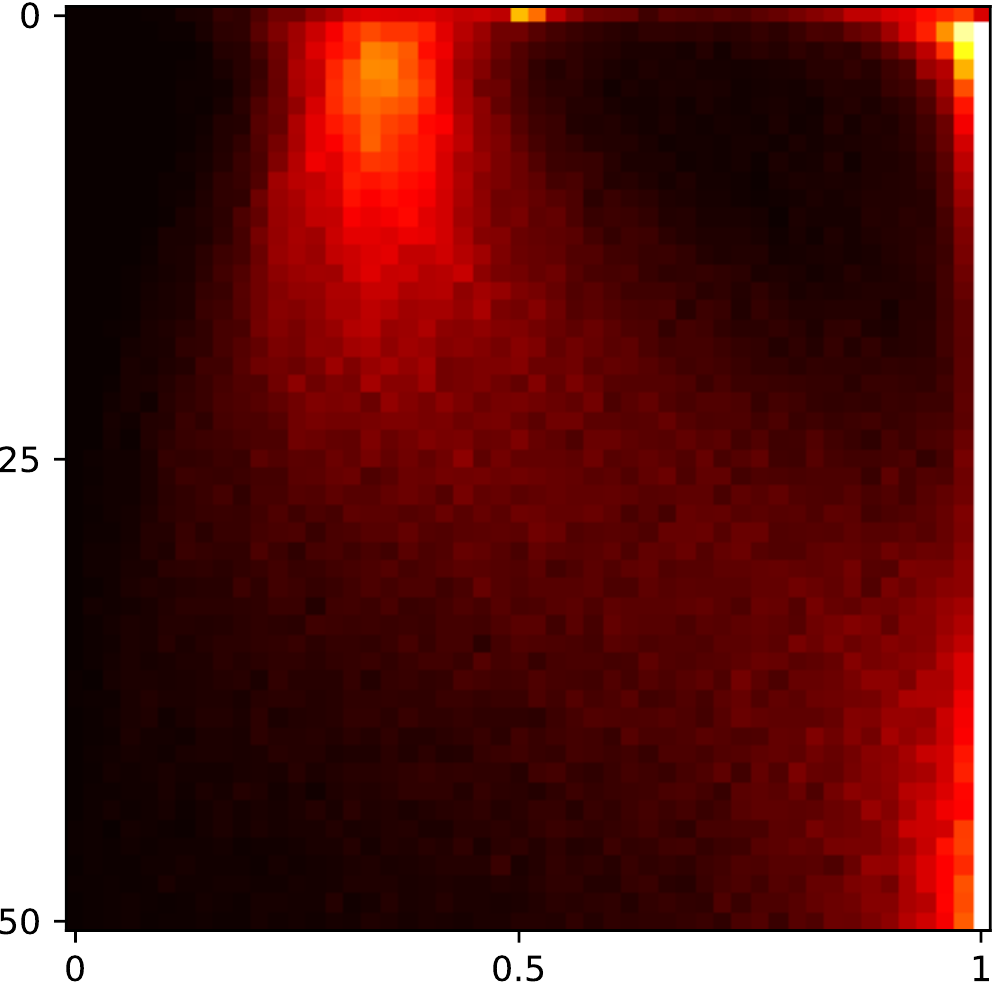}}
  \centerline{(e) No norm+Dropout}\medskip
\end{minipage}
\begin{minipage}[c]{0.24\linewidth}
  \centering
  \centerline{\includegraphics[width=\linewidth]{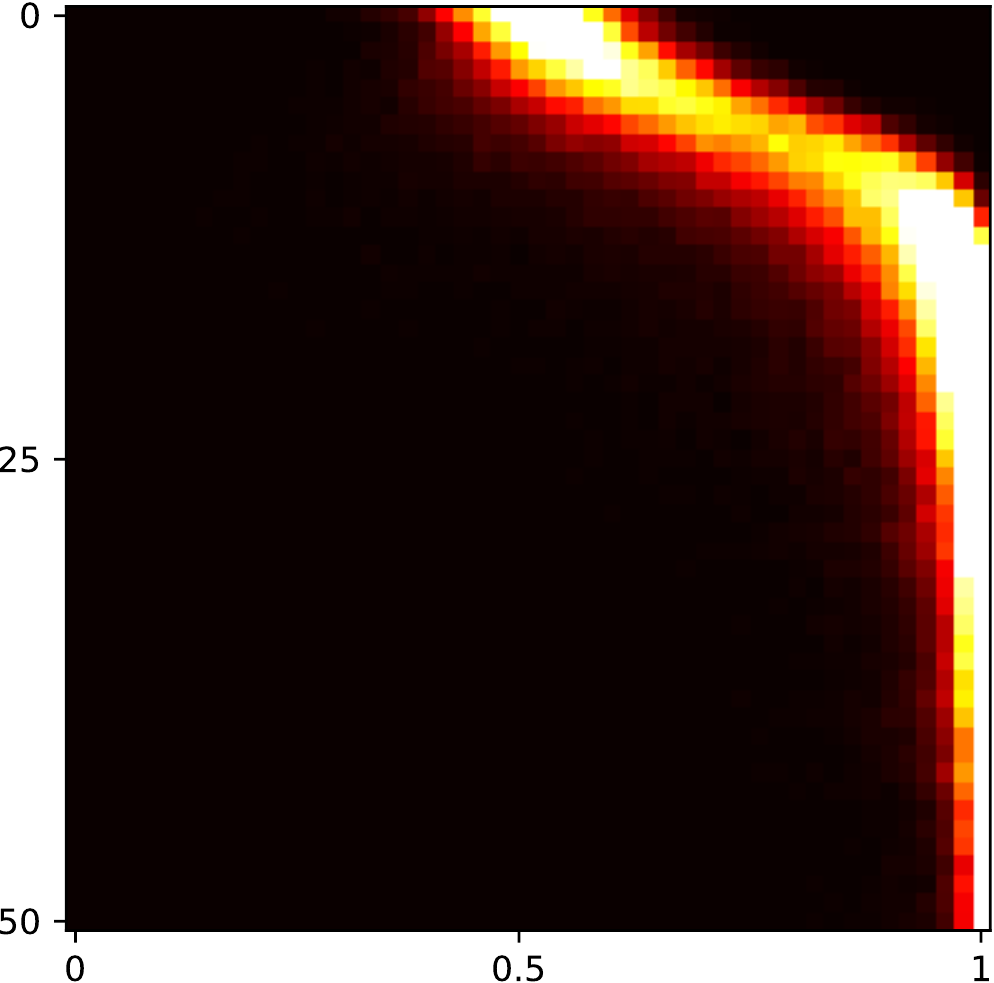}}
  \centerline{(f) BN+Dropout}\medskip
\end{minipage}
\begin{minipage}[c]{0.24\linewidth}
  \centering
  \centerline{\includegraphics[width=\linewidth]{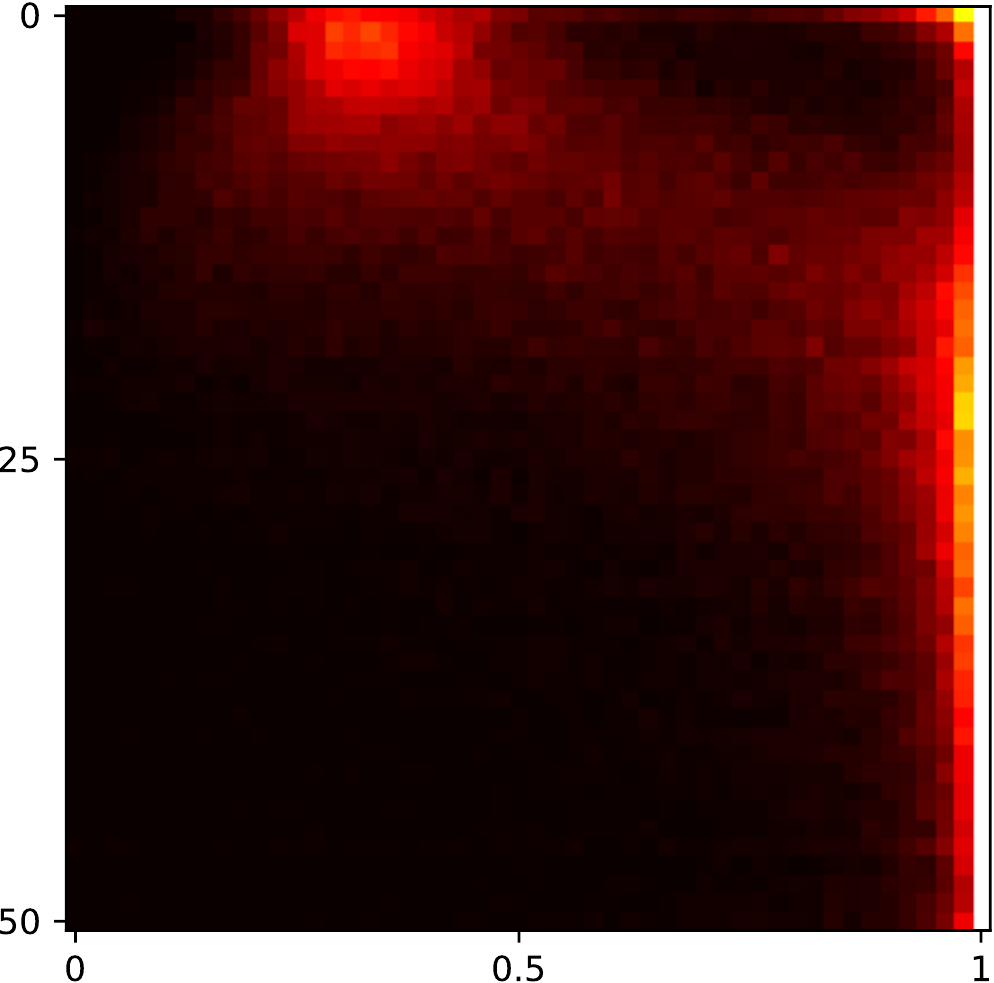}}
  \centerline{(g) LN+Dropout}\medskip
\end{minipage}
\begin{minipage}[c]{0.24\linewidth}
  \centering
  \centerline{\includegraphics[width=\linewidth]{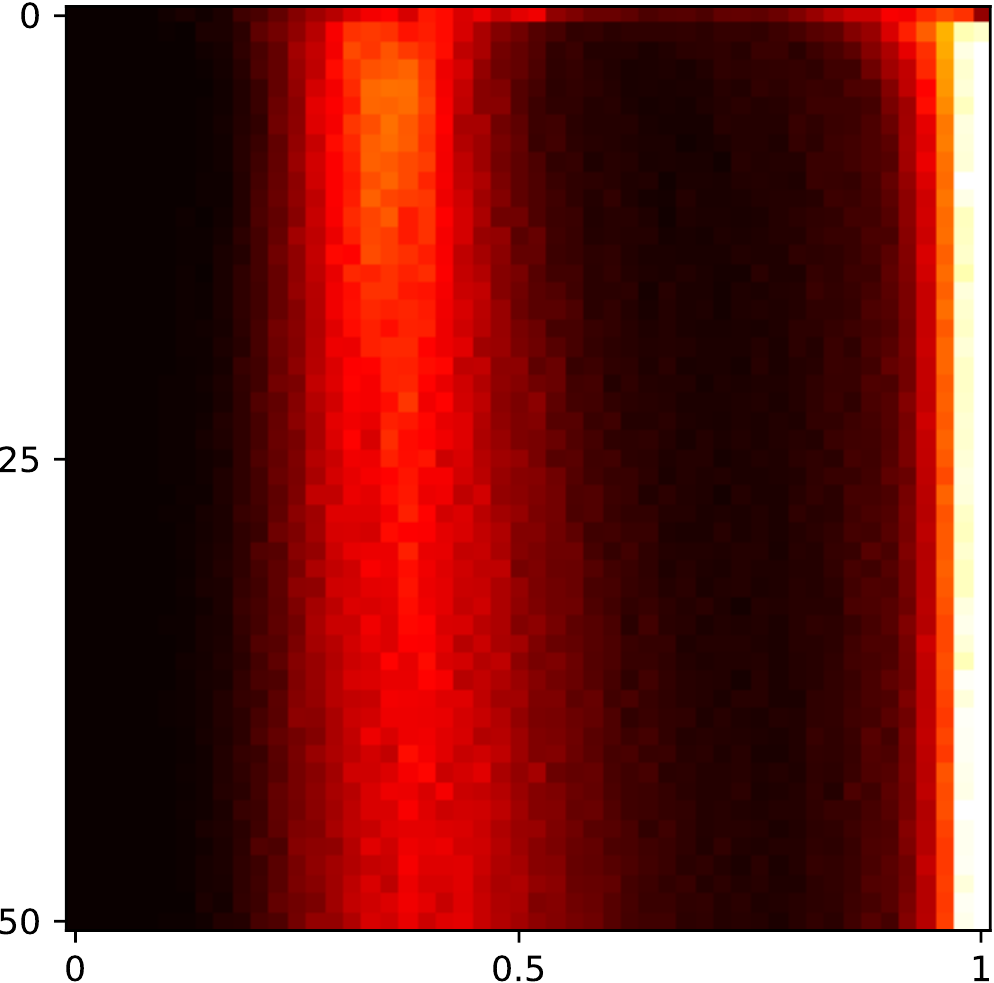}}
  \centerline{(h) SN+Dropout}\medskip
\end{minipage}
\begin{minipage}[c]{0.49\linewidth}
  \centering
  \centerline{\includegraphics[width=\linewidth]{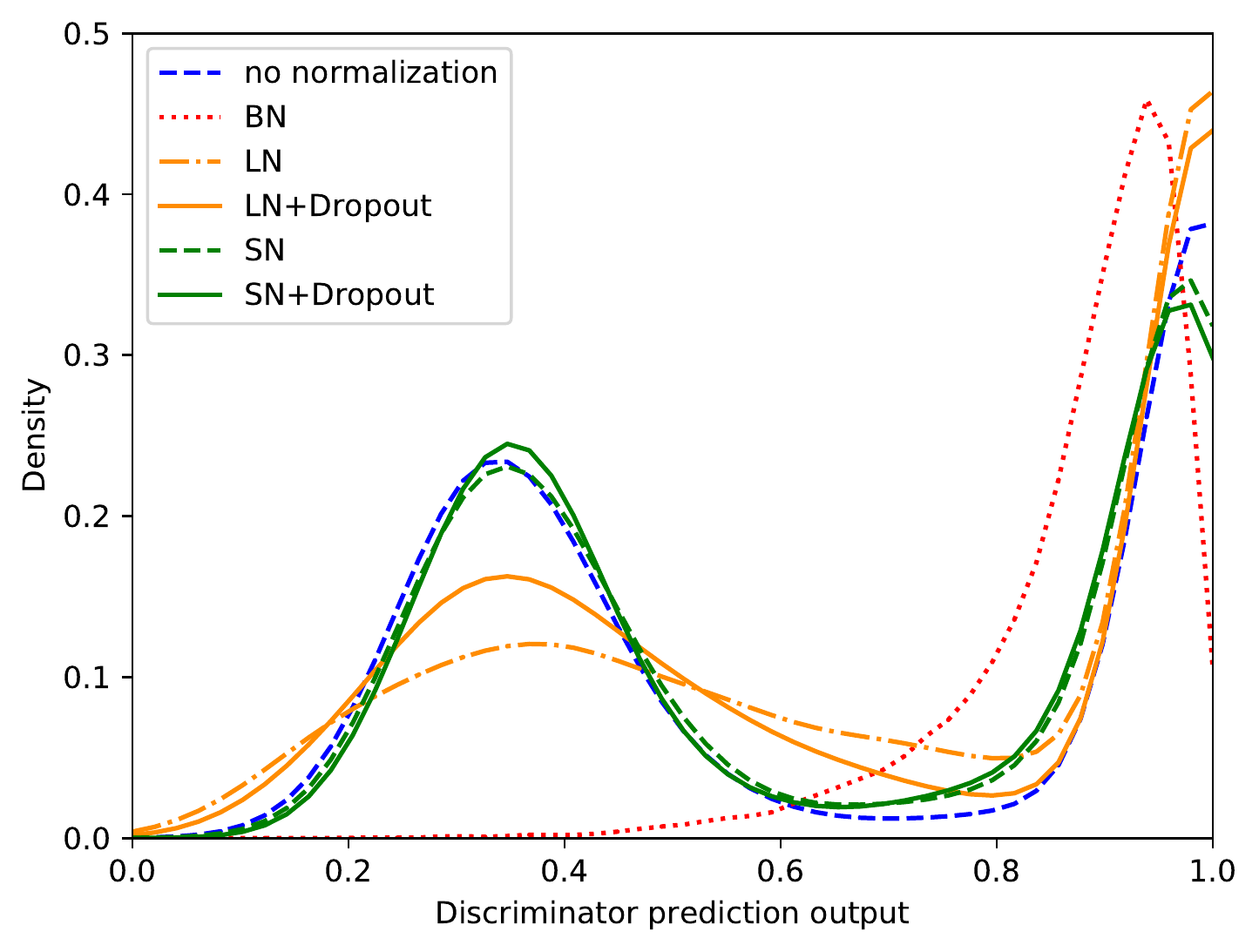}}
  \centerline{(i) Slice after 5 epochs}\medskip
\end{minipage}
\begin{minipage}[c]{0.49\linewidth}
  \centering
  \centerline{\includegraphics[width=\linewidth]{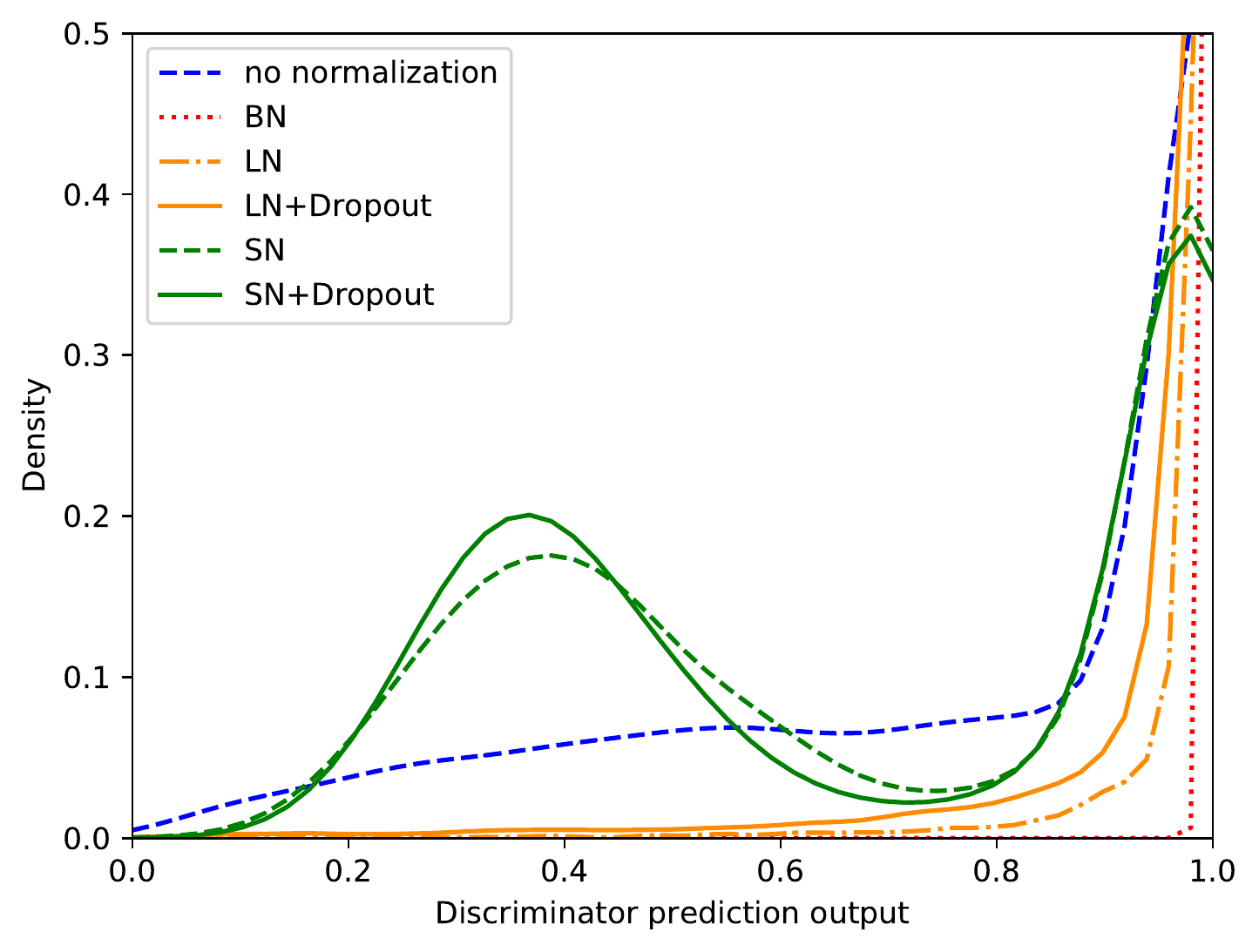}}
  \centerline{(j) Slice after 25 epochs}\medskip
\end{minipage}
\caption{D predictions on unlabeled training examples. (a), (b), (c), (d), (e), (f), (g), (h) images show the evolutions of the histograms of predictions during the training of $D$. Each horizontal line of pixels represents the histogram of predictions, between $0$ and $1$ along the horizontal axis, of $D$ on the entire unlabeled training dataset. Clear hot colors represent a high density of prediction. The vertical axis indicates the training iterations from $0$ to $50$ epochs. Figures (i) and (j) represent the corresponding histograms of predictions after 5 and 25 epochs. Settings are with positive class $8$ and negative class $3$ of MNIST dataset, with $\pi_P = 0.5$.}
\label{fig:Norm_tests_on_D}
\end{center}
%
\end{figure}

We compare in Figure \ref{fig:Norm_tests_on_D} the ability of $D$ to distinguish positive from negative samples distributions included inside the unlabeled training dataset when $D$ is trained on a PU image dataset without normalization and with BN, LN, and SN normalizations. We also consider the cases when they are combined with the dropout regularization. In this experiment, $D$ is trained alone such that it is not adversarially trained with $G$. This enables to better observe and anticipate the adversarial behaviour of $D$, and consequently the behaviour of $G$ during the adversarial training.


We show in Figure \ref{fig:Norm_tests_on_D} the histograms of $D$ predictions concerning the unabeled training examples. As previously explained in the Section \ref{subsec:Biased_PU_risk}, if $D$ associates exclusively the label $1$ with the distribution $p_N$, then we can observe a mixture of two distributions in the corresponding histograms. The one on the right corresponds to $D$ predictions for unlabeled negative examples. The second one on the left corresponds to $D$ predictions for unlabeled positive examples. It is shifted away from the label $1$ and centered around $\delta$. Both distributions cannot be observed with BN. With LN, we can observe both distributions at the beginning of the training before the appearance of an overfitting problem for the unlabeled positive examples. Consequently, at the end of the training, both distributions have merged as with BN. In contrast, SN considerably decreases this overfitting problem. Moreover, the addition of the dropout further helps, such that the dispersion of $D$ predictions is attenuated. This confirms that BN is not compatible with the proposed framework. LN can be used for relatively short trainings. And we conclude that the combination $SN+Dropout$ is the best solution to preserve the distinction between $p_P$ and $p_N$ for long trainings. This is consistent with the arguments discussed in Sec. \ref{subsec:Discriminator_regularizations}.

%

%
%
%
%

Now that we have validated the discriminator ability to separate positive and negative distributions from a positive unlabeled dataset, we select the most appropriate regularization techniques SN and dropout to train adversarially the discriminator and the generator hereafter. The proposed GAN based PU model ability to generate relevant counter-examples is assessed in the next section.

\subsubsection{Generating counter-examples}
\label{subsubsec:counter_examples_generation}

\begin{figure}[!ht]
\begin{center}
\begin{minipage}[c]{0.3\linewidth}
  \centering
  \centerline{Positive}\medskip
\end{minipage}
\begin{minipage}[c]{0.3\linewidth}
  \centering
  \centerline{Unlabeled}\medskip
\end{minipage}
\begin{minipage}[c]{0.3\linewidth}
  \centering
  \centerline{Generated}\medskip
\end{minipage}
\begin{minipage}[c]{0.3\linewidth}
  \centering
  \centerline{\includegraphics[width=\linewidth]{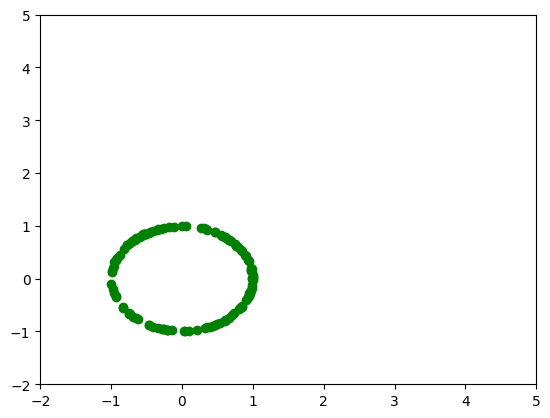}}
  \centerline{(a)}\medskip
\end{minipage}
\begin{minipage}[c]{0.3\linewidth}
  \centering
  \centerline{\includegraphics[width=\linewidth]{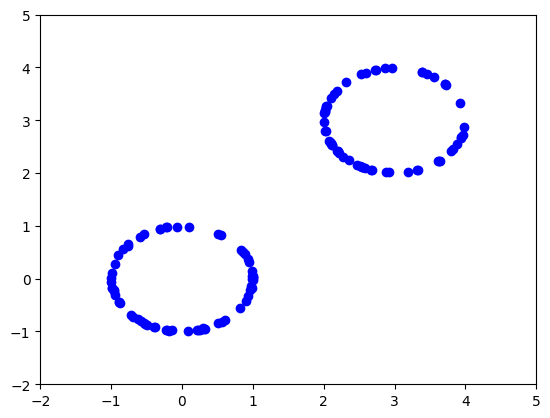}}
  \centerline{(b)}\medskip
\end{minipage}
\begin{minipage}[c]{0.3\linewidth}
  \centering
  \centerline{\includegraphics[width=\linewidth]{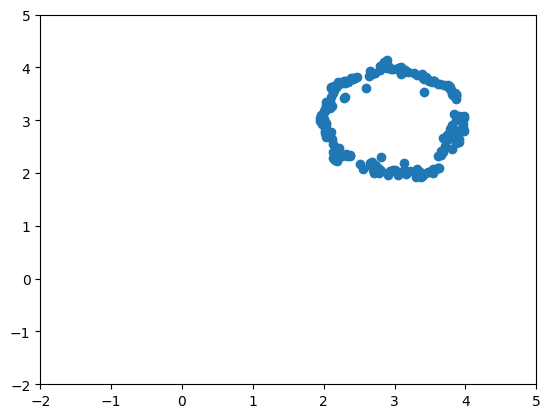}}
  \centerline{(c)}\medskip
\end{minipage}
\begin{minipage}[c]{0.3\linewidth}
  \centering
  \centerline{\includegraphics[width=\linewidth]{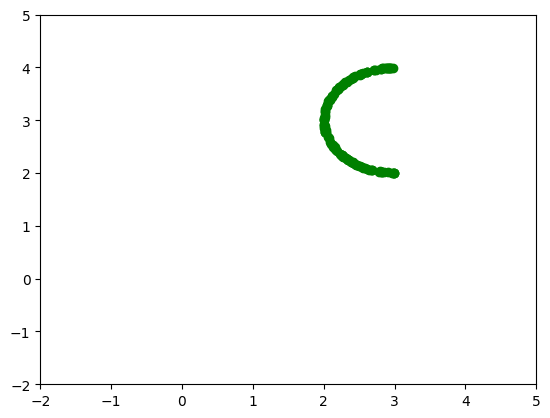}}
  \centerline{(d)}\medskip
\end{minipage}
\begin{minipage}[c]{0.3\linewidth}
  \centering
  \centerline{\includegraphics[width=\linewidth]{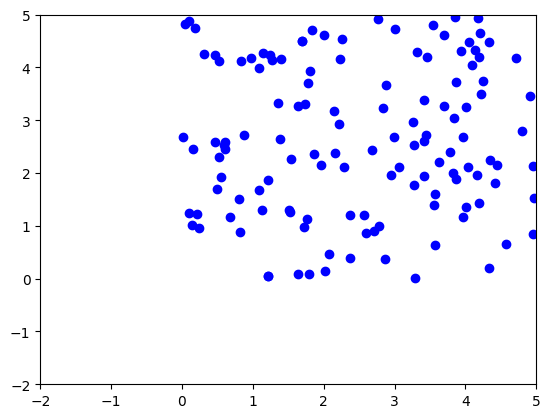}}
  \centerline{(e)}\medskip
\end{minipage}
\begin{minipage}[c]{0.3\linewidth}
  \centering
  \centerline{\includegraphics[width=\linewidth]{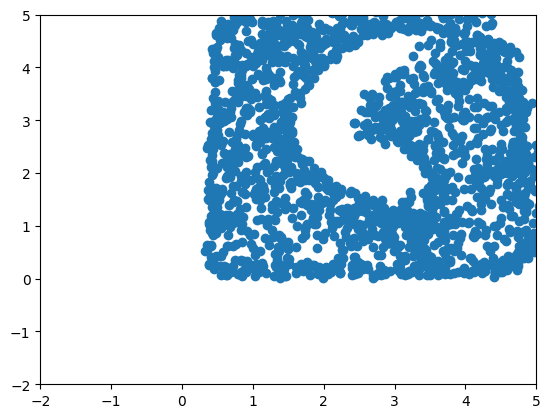}}
  \centerline{(f)}\medskip
\end{minipage}
%

\caption{Proposed approach applied to two different clusters of 2D points. $D$ and $G$ have a multilayer-perceptron structure with respectively $128$ hidden units. From left to right, figures are respectively labeled positive, unlabeled with $\pi_P=0.5$, and generated samples. Figures (a), (b), (c) case is with distributions following circle shapes. Figures (d), (e), (f) case is with a half circle distribution for positive examples, and a uniform distribution over a defined interval for unlabeled examples.}
\label{fig:PU2Dpointclouds}
\end{center}
%
\end{figure}

\begin{figure}[!ht]
\begin{center}
\begin{minipage}[c]{0.10\linewidth}
  \centering
  \centerline{     }\medskip
\end{minipage}
\begin{minipage}[c]{0.29\linewidth}
  \centering
  \centerline{Positive}\medskip
\end{minipage}
\begin{minipage}[c]{0.29\linewidth}
  \centering
  \centerline{Unlabeled}\medskip
\end{minipage}
\begin{minipage}[c]{0.29\linewidth}
  \centering
  \centerline{Generated}\medskip
\end{minipage}

\begin{minipage}[c]{0.10\linewidth}
  \centering
  \centerline{MNIST}
  \centerline{($\pi_P=0.5$) }\medskip
\end{minipage}
\begin{minipage}[c]{0.29\linewidth}
  \centering
  \centerline{\includegraphics[width=\linewidth]{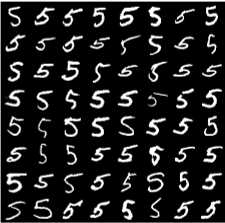}}
  \medskip
\end{minipage}
\begin{minipage}[c]{0.29\linewidth}
  \centering
  \centerline{\includegraphics[width=\linewidth]{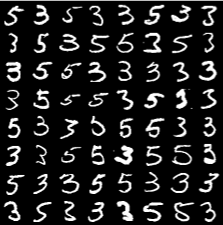}}
  \medskip
\end{minipage}
\begin{minipage}[c]{0.29\linewidth}
  \centering
  \centerline{\includegraphics[width=\linewidth]{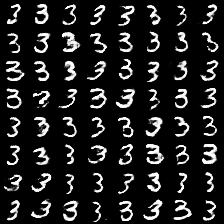}}
  \medskip
\end{minipage}

\begin{minipage}[c]{0.10\linewidth}
  \centering
  \centerline{CIFAR-10}
  \centerline{($\pi_P=0.3$) }\medskip
\end{minipage}
\begin{minipage}[c]{0.29\linewidth}
  \centering
  \centerline{\includegraphics[width=\linewidth]{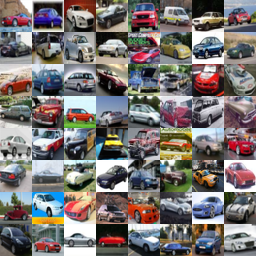}}
  \medskip
\end{minipage}
\begin{minipage}[c]{0.29\linewidth}
  \centering
  \centerline{\includegraphics[width=\linewidth]{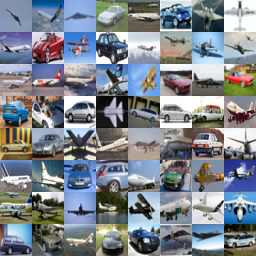}}
  \medskip
\end{minipage}
\begin{minipage}[c]{0.29\linewidth}
  \centering
  \centerline{\includegraphics[width=\linewidth]{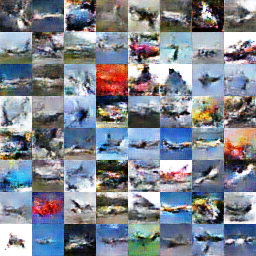}}
  \medskip
\end{minipage}

\begin{minipage}[c]{0.10\linewidth}
  \centering
  \centerline{celebA}
  \centerline{($\pi_P=0.5$) }\medskip
\end{minipage}
\begin{minipage}[c]{0.29\linewidth}
  \centering
  \centerline{\includegraphics[width=\linewidth]{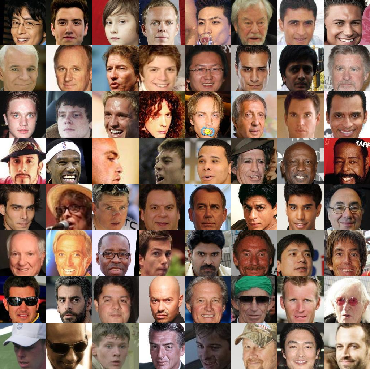}}
  \medskip
\end{minipage}
\begin{minipage}[c]{0.29\linewidth}
  \centering
  \centerline{\includegraphics[width=\linewidth]{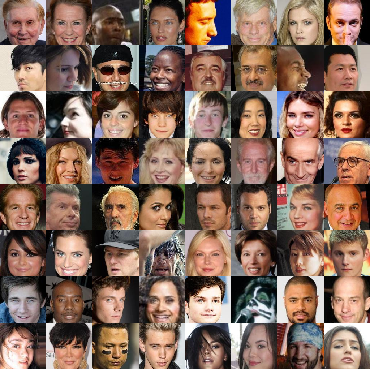}}
  \medskip
\end{minipage}
\begin{minipage}[c]{0.29\linewidth}
  \centering
  \centerline{\includegraphics[width=\linewidth]{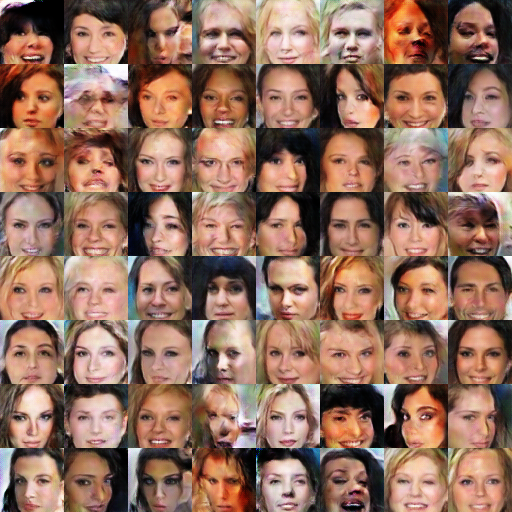}}
  \medskip
\end{minipage}
\caption{Counter-examples generation from Positive Unlabeled image datasets. The two left columns present input positive and unlabeled training samples $x_P$ and $x_U$. The right column presents output generated minibatch samples $x_G$. The first row presents results for MNIST classification task \textit{$5$-vs-$3$} when $\pi_P=0.5$. The second row presents results for CIFAR-10 classification task \textit{Car-vs-Airplane} when $\pi_P=0.3$. The third row presents results for the arbitrary celebA classification task \textit{Male-vs-Female} when $\pi_P=0.5$. Visually, every generated samples observed hallucinate counter-examples included in the unlabeled training set.}
\label{fig:three_data}
\end{center}
%
\end{figure}

From a qualitative point of view, and contrary to the PGAN model, the proposed D-GAN paradigm generates items which only follow the counter-examples distribution for diverse data types. This is illustrated in Figure \ref{fig:PU2Dpointclouds} for 2D point datasets and in Figure \ref{fig:three_data} for image datasets.

In Figure \ref{fig:PU2Dpointclouds}, we can observe on the top line that the generated sample exclusively follows the distribution of the counter-examples included in the unlabeled set (i.e. simultaneously not positive and unlabeled).
On the bottom line, we can observe that the generator has learned the distribution of confident complements of the positive sample distribution over the uniform distribution of unlabeled sample. In addition, we can also observe that a small area around the positive sample distribution is not captured by the generator. This shows the ability of the proposed generative model to not overfit the positive sample distribution boundary.


In Figure \ref{fig:three_data}, we can also observe that the generated examples systematically follow the counter-examples distribution on three image datasets: MNIST, CIFAR-10 and celebA. 

In order to enable reproducibility, a D-GAN implementation corresponding to Figure \ref{fig:celebA_Norm_tests} results is available\footnote{The code is available in supplementary material.} and is applied on the LS-GAN model \citep{qi_loss_sensitive_2017}. Our code also includes the method proposed by \citep{chiaroni_learning_2018} to establish a PU training dataset from a fully labeled dataset with parameters $\rho$ and $\pi_P$.
%
%

Morevover, as mentioned previously, the regularization technique used in the discriminator has a direct impact on the samples generated by the generator. Figure \ref{fig:celebA_Norm_tests} shows samples generated by $G$ depending on the normalization technique used in $D$. We can observe that in the first row, with $\pi_P=0.3$, we naturally obtain around thirty percent of men faces generated using any normalization techniques with the orginial GAN framework used in PGAN. The generated images quality seems visually equivalent between BN, LN or SN. As previously explained, in the second row, also with $\pi_P=0.3$, the proposed D-GAN approach is not compatible with BN. On the contrary, with LN, it exclusively generates counter-examples: women faces with only few men patterns like facial hairs. Finally, it exclusively generates women faces with SN. Those results are consistent with Sec. \ref{subsec:Discriminator_regularizations} and \ref{subsubsec:Without_BN_tests}. The D-GAN trained with $\pi_P=0.5$ and BN naturally generates around fifty percents of men faces, as we recall that BN does not enable to capture the counter-examples distribution. The D-GAN also performs relatively well with SN+Dropout when $\pi_P=0.5$. It exclusively generates women faces. This confirms that the generator behaviour is highly dependent on the discriminator generalization ability, which in turn depends on normalization techniques used. This also confirms that the proposed D-GAN framework presents the interesting ability to exclusively hallucinate counter-examples on a real PU image dataset when it is combined with appropriate discriminator regularizations.

\begin{figure}[!ht]
\begin{center}
\begin{minipage}[c]{0.10\linewidth}
  \centering
  \centerline{     }\medskip
\end{minipage}
\begin{minipage}[c]{0.29\linewidth}
  \centering
  \centerline{BN}\medskip
\end{minipage}
\begin{minipage}[c]{0.29\linewidth}
  \centering
  \centerline{LN}\medskip
\end{minipage}
\begin{minipage}[c]{0.29\linewidth}
  \centering
  \centerline{SN}\medskip
\end{minipage}

\begin{minipage}[c]{0.10\linewidth}
  \centering
  \centerline{GAN     }
  \centerline{($\pi_P=0.3$) }\medskip
\end{minipage}
\begin{minipage}[c]{0.29\linewidth}
  \centering
  \centerline{\includegraphics[width=\linewidth]{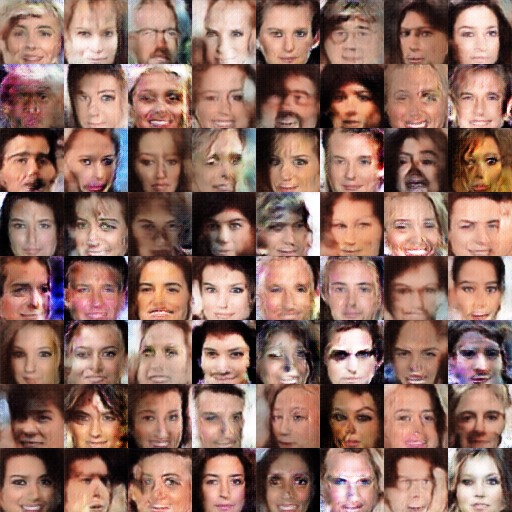}}
  \medskip
\end{minipage}
\begin{minipage}[c]{0.29\linewidth}
  \centering
  \centerline{\includegraphics[width=\linewidth]{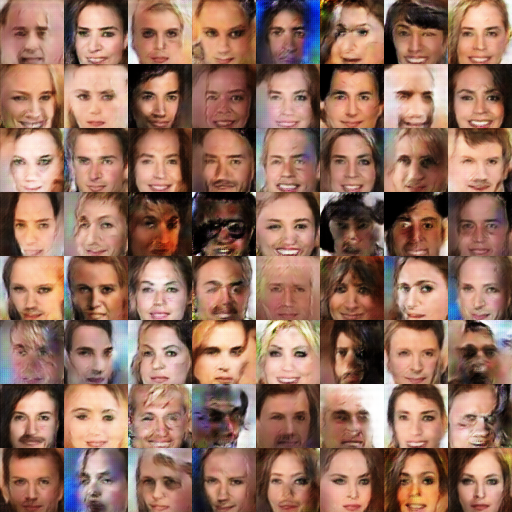}}
  \medskip
\end{minipage}
\begin{minipage}[c]{0.29\linewidth}
  \centering
  \centerline{\includegraphics[width=\linewidth]{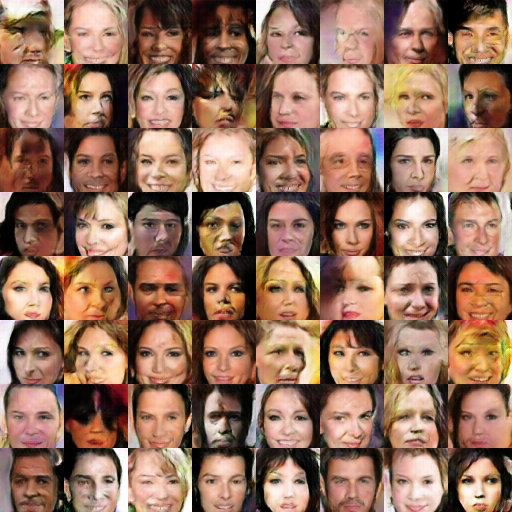}}
  \medskip
\end{minipage}

\begin{minipage}[c]{0.10\linewidth}
  \centering
  \centerline{D-GAN}
  \centerline{($\pi_P=0.3$) }\medskip
\end{minipage}
\begin{minipage}[c]{0.29\linewidth}
  \centering
  \centerline{\includegraphics[width=\linewidth]{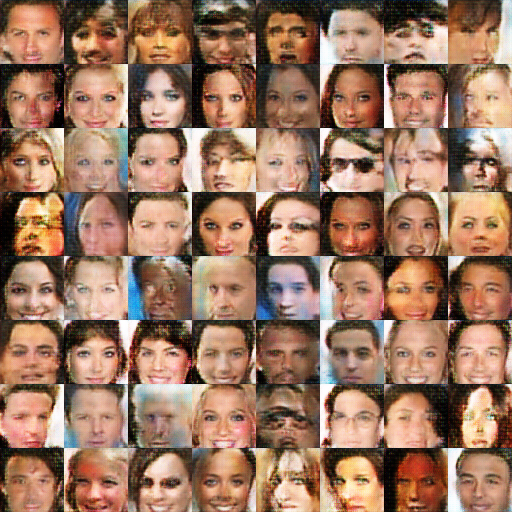}}
  \medskip
\end{minipage}
\begin{minipage}[c]{0.29\linewidth}
  \centering
  \centerline{\includegraphics[width=\linewidth]{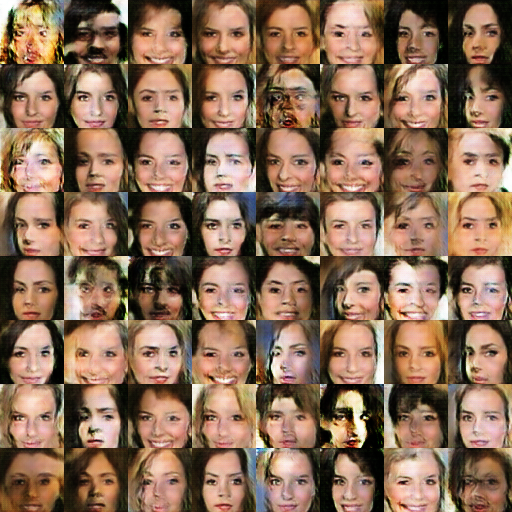}}
  \medskip
\end{minipage}
\begin{minipage}[c]{0.29\linewidth}
  \centering
  \centerline{\includegraphics[width=\linewidth]{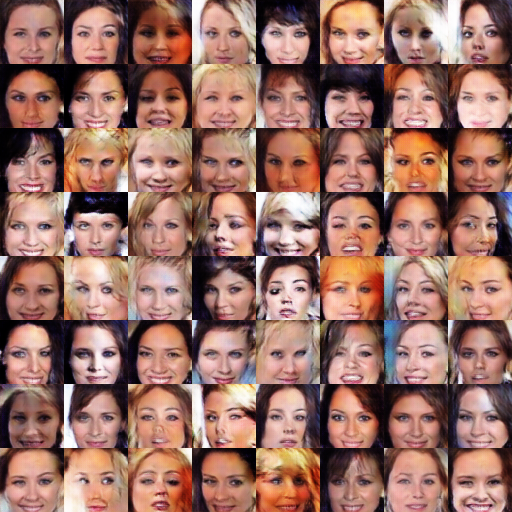}}
  \medskip
\end{minipage}

\begin{minipage}[c]{0.10\linewidth}
  \centering
  \centerline{D-GAN}
  \centerline{($\pi_P=0.5$) }\medskip
\end{minipage}
\begin{minipage}[c]{0.29\linewidth}
  \centering
  \centerline{\includegraphics[width=\linewidth]{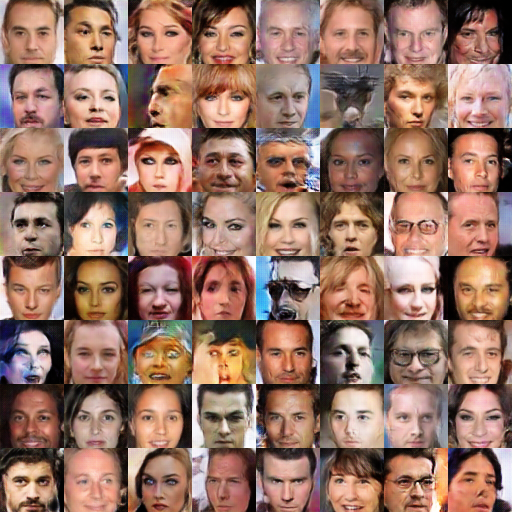}}
  \medskip
\end{minipage}
\begin{minipage}[c]{0.29\linewidth}
  \centering
  \centerline{\includegraphics[width=\linewidth]{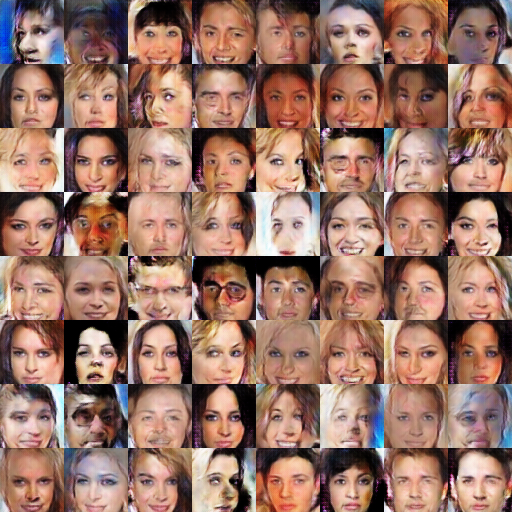}}
  \medskip
\end{minipage}
\begin{minipage}[c]{0.29\linewidth}
  \centering
  \centerline{\includegraphics[width=\linewidth]{illustrations/DGAN_pi_05_SN_Dropout_train_99_0966}}
  \medskip
\end{minipage}
\caption{Discriminator regularizations impacts on the generated samples from a PU celebA image dataset after 100 training epochs iterations. The three columns correspond respectively to training experiments with BN, LN, and SN normalization techniques. The first row presents samples generated using the original LS-GAN discriminator loss function. The two bottom rows present the samples generated by integrating the proposed model discriminator loss function term $\mathbb{E}_{x_P \sim p_P} [MSE(D(x_P),0)]$ in the original LS-GAN loss function, with $MSE$ the mean squared error metric.}
\label{fig:celebA_Norm_tests}
\end{center}
%
\end{figure}

We have shown in this section, from a qualitative point of the view, the discriminator ability to separate positive and negative distributions from a positive unlabeled dataset, and the generator ability to learn the counter-examples distribution on various datasets during the first stage. Next, we propose in Sec. \ref{subsec:D-GAN_versus_state_of_the_art} to quantitatively evaluate the proposed model through an empirical study by focusing on the second-stage classifier $C$ output predictions.

\subsection{Divergent-GAN for Positive Unlabeled learning}
\label{subsec:D-GAN_versus_state_of_the_art}

In this section, we evaluate empirically our method on standard PU learning tasks such that we can test its ability to address respective issues of the state-of-the-art methods presented in Section \ref{sec:Related_work}.

Concerning these comparative experiments, we use the DCGAN \citep{radford_unsupervised_2015} architecture. 

\subsubsection{Robustness to prior noise}
\label{subsubsec:Prior_noise_insensitivity}

Nowadays, the \textit{stochastic gradient descent} (SGD) method remains a useful deep learning regularization technique for large-scale machine learning problems \citep{bottou_large_scale_2010}. SGD provides a regularizing effect by using minibatches \citep{wilson_general_2003}. However, a smaller batch size implies a higher prior noise per batch. Thus, in this section, we empirically study the proposed model robustness to prior noise using small batch sizes.

\begin{figure}[h]
\begin{center}
\begin{minipage}[c]{0.49\linewidth}
  \centering
  \centerline{\resizebox{9.00cm}{!}{\includegraphics[width=\linewidth]{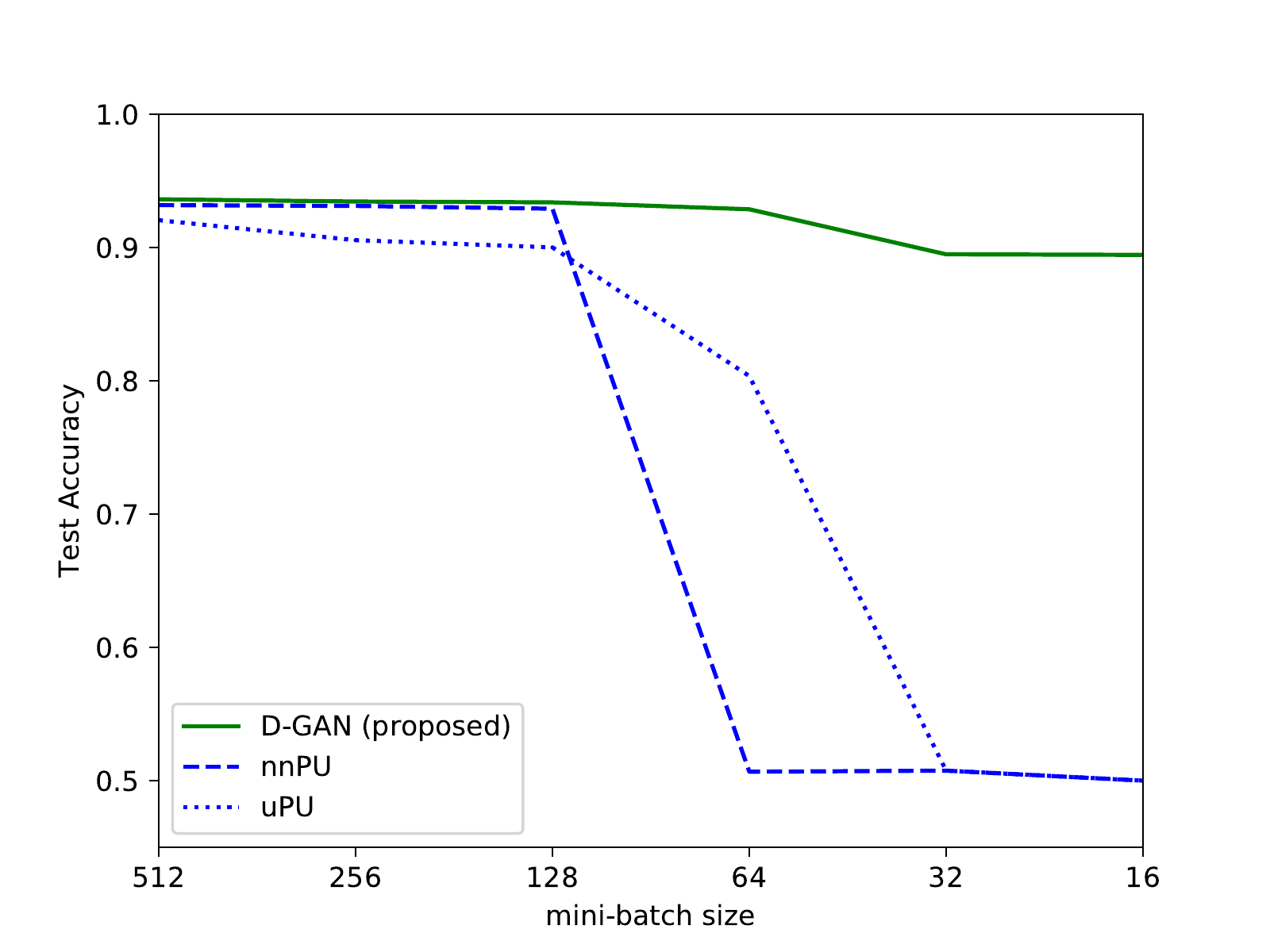}}}
  \centerline{(a) Curves}\medskip
\end{minipage}
\begin{minipage}[c]{0.49\linewidth}
  \centering
  \centering{
  	\resizebox{7.75cm}{!}{
  	\begin{tabular}{cc|ccc}
        \toprule
        \multicolumn{2}{c|}{Even-vs-Odd (MNIST)} & \multicolumn{1}{c}{\textbf{D-GAN}} & \multicolumn{1}{c}{nnPU} & \multicolumn{1}{c}{uPU} \\
        \midrule
        \multicolumn{2}{c|}{Without prior} & $\checkmark$ & $\times$ & $\times$ \\
        \midrule
        \multicolumn{1}{c|}{minibatch size} & $std({\pi_P})\cdot 10^2$ & \multicolumn{3}{c}{Test Accuracy} \\
        \midrule
        \multicolumn{1}{c|}{512} & 2.22 & \textbf{0.936} & 0.932 & 0.921 \\
        \multicolumn{1}{c|}{256} & 3.2 & \textbf{0.935} & 0.931 & 0.906 \\
        \multicolumn{1}{c|}{128} & 4.31 & \textbf{0.934} & 0.929 & 0.9 \\
        \multicolumn{1}{c|}{64} & 6.51 & \textbf{0.929} & 0.507 & 0.804 \\
        \multicolumn{1}{c|}{32} & 8.62 & \textbf{0.895} & 0.508 & 0.508 \\
        \multicolumn{1}{c|}{16} & 13.06 & \textbf{0.907} & 0.5 & 0.5 \\ 
        \bottomrule
    \end{tabular}}}
  \centerline{(b) Detailed scores}\medskip
\end{minipage}
\caption{Prediction test Accuracy on MNIST for the \textit{Even-vs-Odd} classification task, as a function of the minibatch size. We choose the prior value $\pi_P=0.5$, as the standard deviation of the real prior per minibatch is the highest in this way (see Fig. \ref{fig:std_prior_in_function_of_batch_size}). This eases to observe the prior sensitivity. We reproduce the experiment \textit{exp-mnist} proposed by nnPU. The PU dataset contains one thousand positive labeled examples, which are \textit{even digits}. The unlabeled set is composed of the entire initial dataset, thus including also the positive labeled ones. $std({\pi_P})$ is the standard deviation of the prior per minibatch. uPU and nnPU results have been obtained with the code provided by the authors of the nnPU article.
(b) details the prediction scores used to plot the curves in (a). 
}
\label{fig:batch_D_GAN_vs_nnPU}
\end{center}
\end{figure}

We reproduce the \textit{Even-vs-Odd} experiment proposed by \citep{kiryo_positive-unlabeled_2017-1} as a function of the batch training size. It consists of learning to discriminate \textit{even} digits \textit{0, 2, 4, 6, 8} from \textit{odd} digits \textit{1, 3, 5, 7, 9}. Concerning the second-stage classifier, we use the \textit{multilayer perceptron} architecture provided by \citep{kiryo_positive-unlabeled_2017-1} \footnote{The code is available at: \url{https://github.com/kiryor/nnPUlearning}.}. We only replace the bottom fully connected layer of the classifier by a convolutional layer, similarly to the generator top layer and discriminator bottom layer in the DCGAN \citep{radford_unsupervised_2015} structure that we use. This avoids compatibility problems between the generator top convolutional layer output and the bottom classifier layer input. Unwanted artifacts in output of GANs MLP structure are slightly different from unwanted artifacts observed in output of GANs convolutional structures.

It turns out that PU approaches using prior such as uPU, nnPU and GenPU make the assumption that the global training dataset prior $\pi_P$ is fixed and known. But in the same PU context, when the minibatch size decreases, the dispersion of $\pi_P$ per minibatch consequently increases. Figure \ref{fig:batch_D_GAN_vs_nnPU} (a) shows that using small batch training sizes causes critical prediction performances collapse issues for unbiased techniques like nnPU and uPU.

On the other hand, our proposed approach without using prior knowledge is drastically less sensitive to this problem: While nnPU and uPU methods become ineffective in terms of test Accuracy (i.e. Accuracy score around 0.5), the D-GAN still provides a prediction test Accuracy of $0.907$ for training minibatches of size $16$ in $D$, $G$ and $C$ to address the \textit{Even-vs-Odd} MNIST superclass classification task, as detailed in Figure \ref{fig:batch_D_GAN_vs_nnPU} (b).

We can conclude that the D-GAN outperforms nnPU and uPU in terms of prediction performances such that it can use minibatches to take advantage of SGD. This capacity is also interesting for incremental learning requirements where only small sample sizes may be managed at each new training iteration. Moreover, recent studies show that it is possible to continually train GANs models \citep{lesort_hal_01951954}.

Now that we have shown that the proposed model is robust to prior noise, we continue the comparative tests with the methods which do not use prior knowledge $\pi_P$ in their training cost-functions to address the PU learning task.
%


%

%

\subsubsection{One versus Rest challenge}
\label{subsubsec:One_vs_Rest_sssec}

We compare in this section the proposed approach with the PGAN and RP methods that we consider as baselines for the PU learning task without prior knowledge. More specifically, we evaluate them on the challenging One-vs-Rest task which consists in trying to distinguish a class from all the other ones. This task is interesting for binary image classification applications where the labeling effort may be exclusively done on the class of interest, the positive class. Another motivation is that One-vs-Rest binary classification brings the tools for multiclass classification \citep{shalev2014understanding}.

\begin{table}[!ht]
\centering
\resizebox{16.75cm}{!}{
\begin{tabular}{c|cc|ccc|cc|ccc}
\toprule
\multicolumn{1}{c|}{One-vs-Rest} & \multicolumn{5}{c|}{$AVG$\textsubscript{MNIST}} & \multicolumn{5}{c}{$AVG$\textsubscript{CIFAR-10}} \\
\midrule
$\pi_P$ & PN & PNGAN & \textbf{D-GAN} & PGAN & RP & PN & PNGAN & \textbf{D-GAN} & PGAN & RP \\
\midrule
0.1 & 0.993 & 0.988 & {\bf 0.989} (0.01)  & 0.965 (0.01) & 0.967 (0.02) & 0.680 & 0.812 & \textbf{0.815} (0.05) & 0.745 (0.08) & 0.622 (0.10) \\ 
0.3 & 0.993 & 0.988 & {\bf 0.983} (0.01) & 0.958 (0.01) & 0.975 (0.02) & 0.680 & 0.812 & \textbf{0.792} (0.05) & 0.760 (0.03) & 0.730 (0.07) \\
0.5 & 0.993 & 0.988 & {\bf 0.971} (0.01) & 0.946 (0.02) & 0.951 (0.04) & 0.680 & 0.812 & \textbf{0.751} (0.04) & 0.748 (0.03) & 0.716 (0.06) \\
0.7 & 0.993 & 0.988 & {\bf 0.938} (0.02) & 0.875 (0.05) & 0.933 (0.07) & 0.680 & 0.812 & \textbf{0.721} (0.04) & 0.702 (0.03) & 0.684 (0.08) \\
\bottomrule
\end{tabular}}
\caption{One-vs-Rest task with two-stage \textbf{PU methods without prior}, as proposed in PGAN \citep{chiaroni_learning_2018}: From a fully labeled PN dataset, we firstly select a fraction $\rho$ of positive labeled examples that we put in the simulated unlabeled set. Then, we add negative labeled examples in the latter to obtain up to a fraction $\pi_P$ of positive examples in this unlabeled set. Compared to nnPU simulation method, this simulation method has the advantage to simultaneously and independently control the number of positive labeled examples to keep, and the fraction $\pi_P$ for the unlabeled set to simulate. \textit{PNGAN} expression represents GAN-based methods reference for the ideal case where $\pi_P=0$, such that we train during the first stage a GAN exclusively over all the initial cleanly labeled counter-examples set. For each dataset and depending on the fraction $\pi_P$, we have tested respectively the ten One-vs-Rest task possibilities and display the corresponding average test F1-score predictions. The standard deviation is indicated in parenthesis.}
\label{tab:meanF1Score}
\end{table}

Table \ref{tab:meanF1Score} shows average predictions for the One-vs-Rest task over MNIST and CIFAR-10 datasets. We use the F1-Score metric for its relevance in such information retrieval and binary classification tasks as highlighted by \citep{bollmann_measurement_theoretical_1980}, \citep{shaw_foundation_1986}, \citep{liu_partially_2002}: the F1-score measures the positive examples retrieval. The PU datasets are simulated as proposed by PGAN such that we can evaluate the results as a function of several $\pi_P$ fractions. Concerning the second-stage classifier in these experiments, we have used the convolutional architecture presented in Figure \ref{fig:archi_MNIST_CIFAR_10} (c). We can observe that the D-GAN globally outperforms PGAN and RP methods in terms of test F1-Score on both MNIST and CIFAR-10 datasets. Moreover, PNGAN results highlight the GAN-based methods data augmentation advantage on complex datasets. This justifies the superior scores obtained by our method compared to RP over the CIFAR-10 dataset.

\begin{figure}[htb!]
\begin{center}
\begin{minipage}[c]{0.49\linewidth}
  \centering
  \centerline{\includegraphics[width=\linewidth]{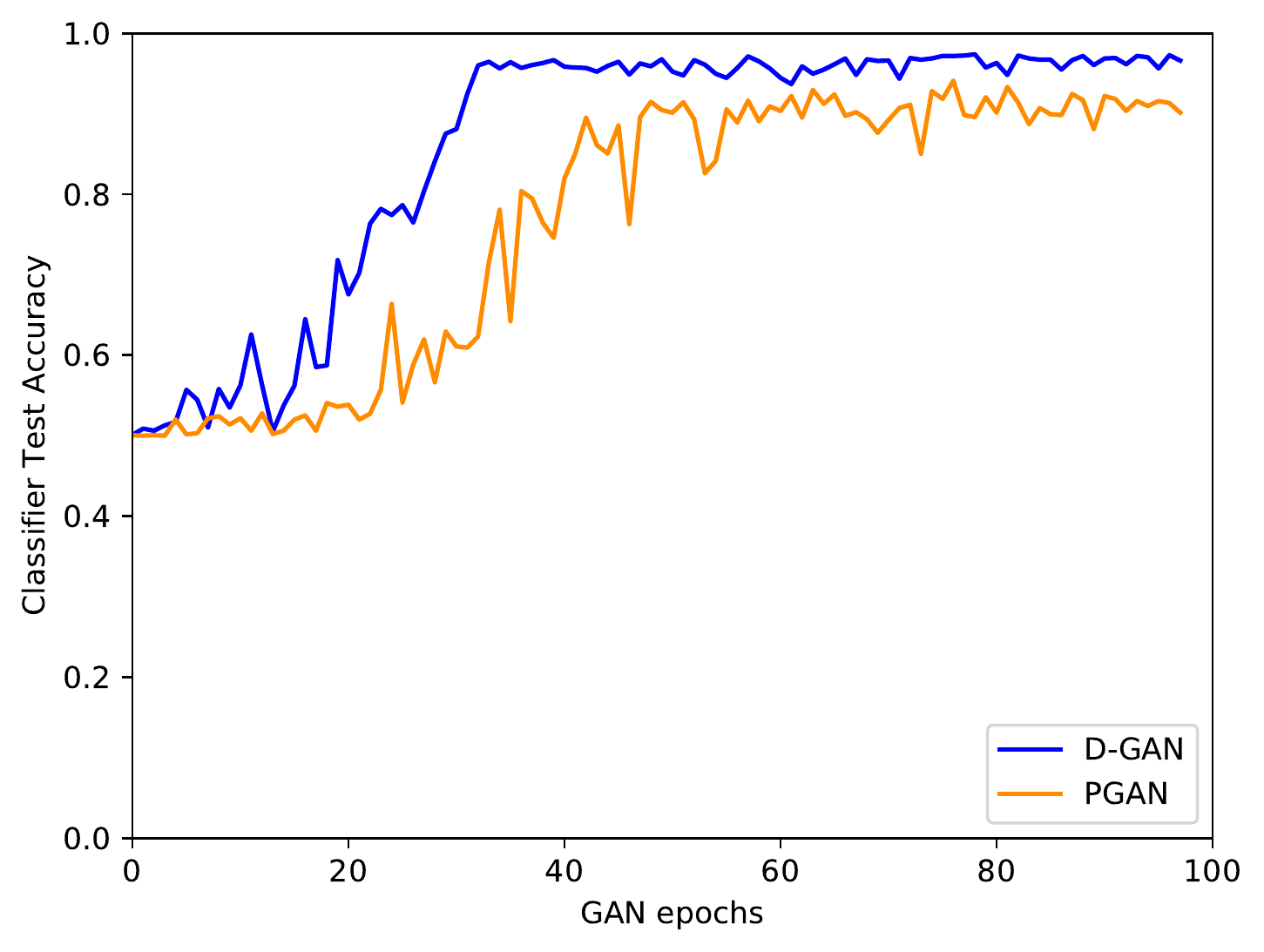}}
  \centerline{(a)}\medskip
\end{minipage}
\begin{minipage}[c]{0.49\linewidth}
  \centering
  \centerline{\includegraphics[width=\linewidth]{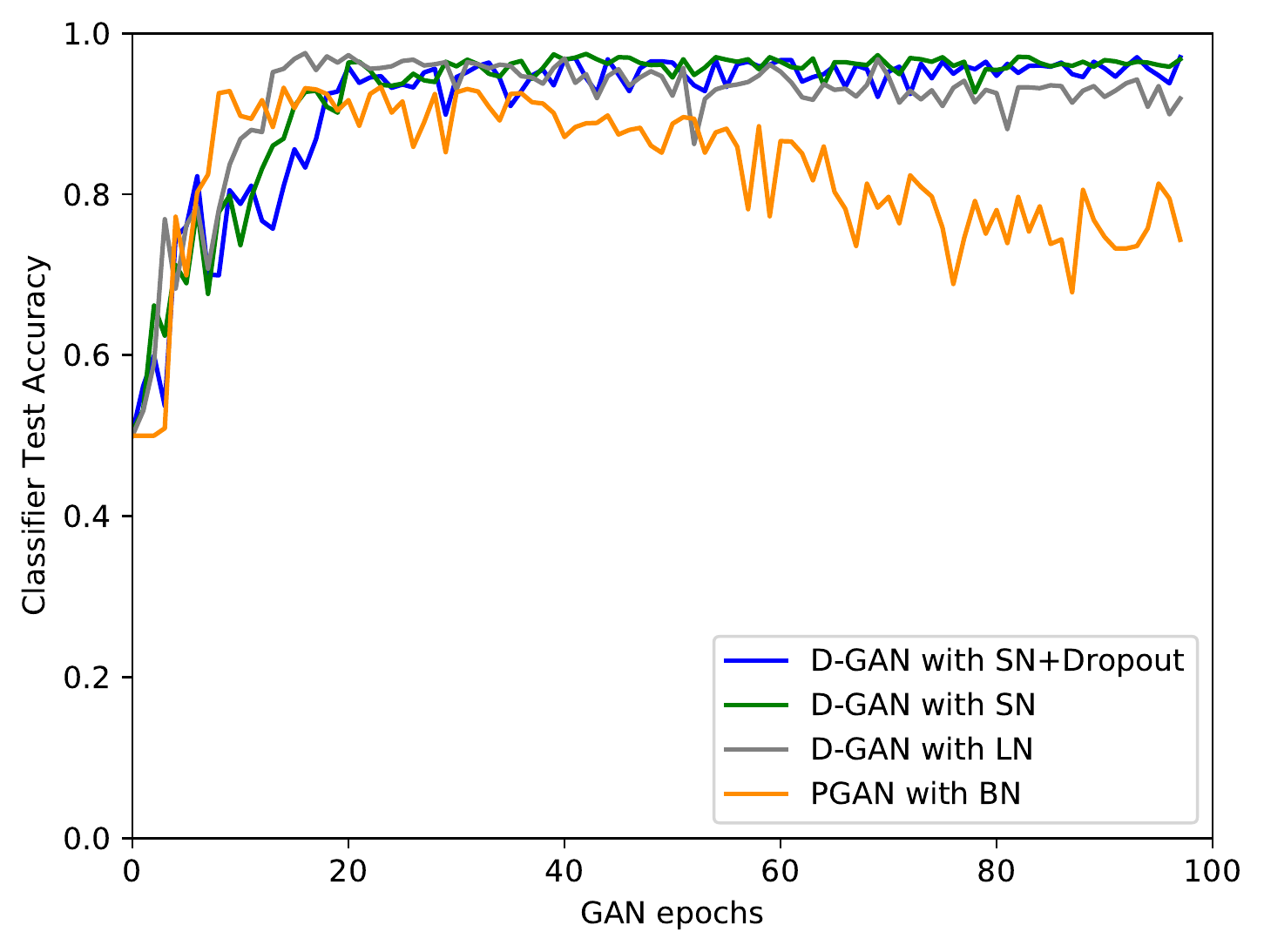}}
  \centerline{(b)}\medskip
\end{minipage}
\caption{Second-stage Classifier (architecture presented in Figure \ref{fig:archi_MNIST_CIFAR_10} (c)) test Accuracy evolution as a function of the first-stage GAN epochs. 8-vs-Rest MNIST task, with $\rho=0.5$ and $\pi_P=0.5$. (a) D-GAN and PGAN are trained without normalization layers. (b) D-GAN and PGAN are respectively trained with LN, SN, SN + dropout, and BN inside the discriminator.}
\label{fig:evo_curves}
\end{center}
%
\end{figure}
%


\textbf{Reducing the overfitting problem:} 
In addition, we can observe that the proposed model also outperforms PGAN on MNIST with a significant margin. This is due to the fact that, compared to the PGAN which is trained to generate unlabeled examples, the proposed approach only generates counter-examples as previously shown in Figures \ref{fig:PU2Dpointclouds} and \ref{fig:three_data}. Consequently, the proposed first-stage generative model does not learn the positive samples distribution, and it avoids the PGAN first-stage overfitting issue on simple datasets like MNIST. Figure \ref{fig:evo_curves} illustrates this phenomenon. In Figure \ref{fig:evo_curves} (a), without normalization, the D-GAN method gets faster a better Accuracy than PGAN when both are trained under the same conditions. In Figure \ref{fig:evo_curves} (b), the D-GAN with LN, SN or SN+dropout follows the learning speed of the PGAN with BN, while demonstrating a steadier behaviour once the Accuracy progression is finished, as it overcomes the PGAN first-stage overfitting problem.

To sum up, in Sec. \ref{subsec:qualitative_analysis}, we demonstrate that the proposed approach is effective at capturing and observing the counter-examples distribution of our class of interest from only positive and unlabeled data, without using the prior information $\pi_P$. In addition, comparative experiments in Sec. \ref{subsec:D-GAN_versus_state_of_the_art} have subsequently highlighted the proposed model ability to address state-of-the-art PU learning issues such as prior sensitivity and first-stage overfitting. It turns out that addressing simultaneously thoses issues fosters the proposed approach to outperform PU state-of-the-art methods in terms of prediction scores without using prior on both simple and complex image datasets.

\section{Conclusion}
\label{sec:Conclusion}

To conclude, we have incorporated into the GAN discriminator loss function a constrained PU risk to deal with PU learning. In this way, the proposed model generates relevant counter-examples from a PU dataset. It outperforms state-of-the-art PU learning methods by addressing their respective issues. Namely, it addresses the prior knowledge dependence of cost-sensitive PU methods and the lack of generalization of selective processes. Moreover, it reduces the overfitting PGAN first-stage problem, while keeping the practical standard GAN architecture, such that it is easily adaptable to recent GANs variants. A side contribution of this article is to have identified discriminator normalizations effects appearing when we manipulate multiple minibatches distributions when dealing with a PU training dataset. 

We believe that the proposed approach stability and prediction performances still have the potential to be improved by taking the best of the representation learning and weakly supervised learning domains. Recent promising GAN training approaches \citep{karras2018progressive}, \citep{zhang_self_attention_2018}, \citep{brock2018large} not mandatorily using BN, may be suitable to extend the proposed approach for higher dimensional image datasets.

\section{References}
\bibliography{mybibfile}


\end{document}